\documentclass[12pt]{article}

\RequirePackage{amsthm,amsmath,amsfonts,amssymb}
\RequirePackage[numbers,sort&compress]{natbib}
\RequirePackage[colorlinks,citecolor=blue,urlcolor=blue]{hyperref}
\RequirePackage{graphicx}
\usepackage{tabularx}
\usepackage{booktabs}
\usepackage{setspace}
\usepackage{authblk} 
\usepackage{hyperref}
\usepackage{url}
\usepackage[top=0.8in, bottom=0.8in, left=0.6in, right=0.6in]{geometry}
\usepackage{algorithm}
\usepackage{algpseudocode}

\floatname{algorithm}{Algorithm}
\usepackage{graphicx}
\usepackage{amsthm}
\theoremstyle{plain}

\theoremstyle{definition}
\newtheorem{definition}{Definition}

\usepackage{thmtools, thm-restate}
\usepackage{xcolor}
\usepackage[font=small,labelfont=bf,width=0.85\textwidth]{caption}

\newcommand{\blind}{1}
\begin{document}

\title{A Graph Sufficiency Perspective for Neural Networks}

\author[1]{Cencheng Shen\thanks{Corresponding author: \href{mailto:shenc@udel.edu}{shenc@udel.edu}}}
\author[2]{Yuexiao Dong}

\affil[1]{Department of Applied Economics and Statistics, University of Delaware, 210 South College Ave, Newark DE 19716, USA}
\affil[2]{Department of Statistics, Operations, and Data Science, Temple University, 1801 N Broad St, Philadelphia PA 19122, USA}
\maketitle

\abstract{This paper analyzes neural networks through graph variables and statistical sufficiency. We interpret neural network layers as graph-based transformations, where neurons act as pairwise functions between inputs and learned anchor points. Within this formulation, we establish conditions under which layer outputs are sufficient for the layer inputs, that is, each layer preserves the conditional distribution of the target variable given the input variable. We explore two theoretical paths under this graph-based view. The first path assumes dense anchor points and shows that asymptotic sufficiency holds in the infinite-width limit and is preserved throughout training. The second path, more aligned with practical architectures, proves exact or approximate sufficiency in finite-width networks by assuming region-separated input distributions and constructing appropriate anchor points. This path can ensure the sufficiency property for an infinite number of layers, and provide error bounds on the optimal loss for both regression and classification tasks using standard neural networks. Our framework covers fully connected layers, general pairwise functions, ReLU and sigmoid activations, and convolutional neural networks. Overall, this work bridges statistical sufficiency, graph-theoretic representations, and deep learning, providing a new statistical understanding of neural networks.}\\

\noindent\textbf{Keywords:} graph anchor points, sufficient transformation, fully connected layer, convolutional layer, region separation

\section{Introduction}

Neural networks have become the dominant framework in modern machine learning, powering breakthroughs in vision and language learning. Classical fully connected feedforward networks, composed of layers of affine transformations and nonlinear activations such as ReLU or sigmoid, form the foundational architecture. Moreover, modern layer architectures have significantly pushed the boundary and achieved huge success, such as the convolutional neural networks (CNNs) \cite{lecun1998gradient, krizhevsky2012imagenet} for image classification and object recognition, and most recently the multi-head self-attention mechanism \cite{vaswani2017attention} for generative AI in large language models \cite{radford2019languagegpt2}. 

The theoretical understanding of neural networks has largely focused on approximation capabilities of classical fully connected networks. Universal approximation theorems \cite{cybenko1989approximation, hornik1989multilayer, barron1993universal} establish that shallow neural networks with sufficient width can approximate any continuous function on compact domains. Recent work has refined these bounds, studying depth-efficiency tradeoffs \cite{lu2017expressive, hanin2019universal} and providing insights into convergence \cite{jacot2018neural} and generalization \cite{neyshabur2017exploring}. For convolutional networks, theoretical work has investigated their compositional properties \cite{cohen2016group, mallat2016understanding}, equivariance to group actions \cite{cohen2016group}, and sparse coding \cite{bruna2013invariant}. 

Despite those results, there is a lack of investigations into the statistical property of neural networks. One such foundational question is: Do neural network layers preserve all information of the input data, i.e., the conditional distribution $P(Y \mid X)$? Much of the literature on universal approximation theorems focuses on whether neural networks can approximate a target function $f(X)$, not whether the neural networks model retain the full information necessary for loss-optimal prediction, nor how transformations of the input space affect the conditional distribution of the target variable. While preserving the target function means preserving the information, retaining information is a more general notion --- namely, the model does not need to approximate the actual functional relationship, as long as all information necessary for loss prediction is retained in the model, then that shall suffice.

To that end, in this work, we aim to understand the statistical sufficiency property of neural network layers via a graph variable framework. In classical statistics, a transformation $Z = T(X)$ is sufficient for the random variable $X$ with respect to the target variable $Y$, if $P(Y \mid Z) = P(Y \mid X)$ almost surely. Therefore, sufficiency ensures that the transformation $T(\cdot)$ preserves all information relevant for predicting $Y$. Specifically, from classical statistical learning literature \cite{hastie2009elements,DevroyeGyorfiLugosiBook}, when sufficiency holds and conditional density is preserved, the optimal loss in either classification or regression is preserved. As such, it is a desirable property for any general transformation.  

We start by adopting a graph-theoretic perspective in a classical fully connected layer. Interpreting the output of a fully connected layer as a generalized similarity graph, each layer computes a pairwise function between the input $X$ and a set of anchor points. Under reasonable assumptions, these anchor points become asymptotically dense in the support of $X$ during training and inference. This setup allows us to establish asymptotic sufficiency of the layer output with respect to the number of neurons $m$. This can be further extended to include special pairwise graph feature including inner products, Euclidean distances, cosine similarity, or kernels. We then include the activation functions into the framework, followed by generalization into multiple fully connected layers. 

To better connect with practical settings, where the number of neurons is inherently finite, we extend the analysis to finite-width networks. We begin with the case of discrete random variables, where finite representations are naturally compatible. We then broaden the analysis to region-separated random variables. For such variables, or those that can be approximated within an error of at most $\epsilon$, we prove that any region-separating transformation is sufficient (or approximately sufficient with the same error bound $\epsilon$). Building on this result, we show that a fully connected layer combined with the ReLU activation can serve as a region-separating transformation, thereby achieving sufficiency (or approximately so with error at most $\epsilon$), provided that each region is convex. Moreover, this sufficiency is preserved as the number of neural network layers increases.

We proceed to quantify error bounds on the optimal prediction loss in both regression and classification settings, when the neural network is approximately sufficient. Finally, we extend the graph sufficiency framework to convolutional layers, interpreting their operations as inner products with regularized anchor points. We show that convolutional layers, too, can act as region-separating transformations and are thus sufficient.

\paragraph*{Paper Structure} We begin with related work in Section~\ref{sec:relatedwork}, which summarizes previous papers and highlights how our method differs. Section~\ref{sec:preliminaries} reviews sufficiency and the classical neural networks in random variables. Section~\ref{sec:graphassumption} shows the graph framework for neural network, as well as introducing anchor points and prove when those anchor points can be dense in the support of input variable.  Section~\ref{sec:singlelayer} presents the asymptotic sufficiency theorem for fully-connected layer, which is then extended to multi-layer in Section~\ref{sec:multilayer}. Section~\ref{sec:region} shows that a region-separating transformation is sufficient for region-separated random variables, as well as showing any random variables can be viewed as $\epsilon$-region-separated with possibly infinite regions. In Section~\ref{sec:regionLayer}, we prove that a fully connected layer can act as a region-separating transformation, thus achieving sufficiency using a finite number of neurons. This section also quantifies the optimal loss difference under approximate sufficiency. Section~\ref{sec:conv} extends the framework to convolutional layers, showing that they too are region-separating, and therefore sufficient, under additional regularity assumptions. Section~\ref{sec:exp} validates the theoretical results via simulations. Section~\ref{sec:conclusion} concludes. All mathematical proofs are provided in the appendix, and Python simulation code is provided in the supplementary. 

\section{Related Work}
\label{sec:relatedwork}

\paragraph*{Neural Network Expressivity and Approximation}
The universal approximation theorem \cite{cybenko1989approximation, hornik1989multilayer} establishes that feedforward networks with a single hidden layer and nonlinear activation can approximate any continuous function on compact subsets of $\mathbb{R}^p$. Subsequent work strengthened this result by quantifying approximation rates \cite{barron1993universal} and analyzing the benefits of depth \cite{lu2017expressive}. ReLU networks, in particular, are known to partition the input space into linear regions, with the number of such regions growing exponentially with depth \cite{montufar2014number}. Neural tangent kernel theory \cite{jacot2018neural} provides insights into the training dynamics and approximation behavior of infinitely wide networks. For convolutional architectures, theoretical studies have examined group equivariance \cite{cohen2016group}, multiscale scattering transforms \cite{bruna2013invariant, mallat2016understanding}, and connections to sparse dictionary learning \cite{papyan2017convolutional}.

\paragraph*{Statistical Sufficiency and Information Bottleneck}
Sufficiency is a classical concept in statistics, foundational to decision theory, causal inference, and sufficient dimension reduction \cite{li1991sliced, cook2005sufficient, lehmann2005testing, li2011principal, li2018sufficient}. It refers to identifying transformations of the input variable that preserve all information relevant to the prediction of the target. For both regression and classification, sufficiency ensures that the optimal predictive loss remains unchanged \cite{hastie2009elements,DevroyeGyorfiLugosiBook}, making it a desirable property for feature learning.

Several recent works have begun to consider sufficiency in the context of neural networks. It was shown that the first layer of neural networks can estimate the central mean subspace in a regression model under smoothness and rank assumptions \cite{xu2025sdr}. Other approaches consider mutual information–based objectives to learn sufficient representations via variational training \cite{zheng2022deep_sufficiency} or contrastive learning \cite{wang2022msmr}. 

A related line of work centers on the information bottleneck principle \cite{tishby2015ib}, which seeks to compress the input representation while retaining as much predictive information as possible. This framework has been extended with rigorous generalization bounds \cite{goldfeld2020ib} and risk guarantees under mutual information constraints \cite{kawaguchi2023ib}. 

\paragraph*{Graph-Based Representations and Anchor Methods}
A key insight of our work is that neural network representations can be interpreted as defining a similarity graph over the input space with anchor points. This connects to classical ideas in manifold learning and semi-supervised learning, where pairwise similarity functions define neighborhood graphs for downstream tasks \cite{belkin2003laplacian, zhou2004learning}. Inner product features correspond to linear graphs, while universal kernels such as the Gaussian kernel yield richer similarity graphs \cite{scholkopf2002learning}.

In graph-structured data, graph embedding has been an important area of research, including node2vec \cite{grover2016node2vec, zhang2024theoretical}, spectral embedding \cite{JMLR:v18:17-448, Priebe2019}, one-hot encoder embedding \cite{GEEOne, GEEPrincipalCommunity}, etc. In particular, graph neural networks have become a widely used architecture \cite{kipf2017semi, Wu2019ACS, tsitsulin2023graph, GEEGCN}. Anchor-based methods have also been proposed to enhance graph representation learning \cite{zhang2023ieagnn}. 

\paragraph*{Contribution of Our Work}
Our paper introduces a population-level framework for analyzing when and how neural network representations preserve sufficiency of the input with respect to the target variable. Unlike the information bottleneck literature, we do not rely on compression or mutual information constraints. Unlike works that explicitly modify the learning algorithm to enforce sufficiency, we analyze the sufficiency properties of standard neural network architectures, including pairwise feature maps, multi-layer networks, and convolutional layers, without altering their training procedures. And unlike prior investigations of sufficient dimension reduction in neural networks, our framework applies to both regression and classification tasks, does not assume a specific functional form or distribution, and addresses the sufficiency of the entire learned representation, not just the central mean subspace.

Our key contribution is a graph-based perspective of neural networks using anchor points, offering two complementary paths to statistical sufficiency: (1) asymptotic injectivity through dense anchor coverage, and (2) exact sufficiency under a region separation condition with appropriately constructed anchors. These results differ significantly from prior works and establish a unique connection between graph-based representations, neural network architectures, and the classical statistical concept of sufficiency.

\section{Preliminaries}
\label{sec:preliminaries}

This section introduces the mathematical basics for sufficient transformation, as well as architectural components for classical fully connected neural networks. We begin by formalizing conditional distributions and sufficiency, then introduce graph-based similarity structures. Then, we review the classical neural network modules on random variables.

\subsection{Conditional Distributions and Sufficiency}

Let $(X, Y)$ be random variables defined on a probability space, with $X \in \mathbb{R}^p$ and $Y \in \mathcal{Y}$ (discrete or continuous). The conditional distribution $P(Y \mid X)$ describes the distribution of $Y$ given $X = x$. Throughout this paper, all random variables are assumed to be Borel measurable and take values in Euclidean space.

\begin{definition}[Sufficiency]
A transformation $T: \mathbb{R}^p \to \mathbb{R}^m$ is said to be sufficient for $X$ with respect to $Y$ if
\[
P(Y \mid T(X)) = P(Y \mid X)
\]
almost surely with respect to the distribution of $X$. Equivalently, $Y \perp X \mid T(X)$.
\end{definition}

Sufficiency implies that $T(X)$ retains all information in $X$ relevant to predicting $Y$. In supervised learning, if a representation $Z = T(X)$ is sufficient, then replacing $X$ with $Z$ does not degrade the optimal loss in either regression or classification. A widely known result is that sufficiency is always preserved under injective transformation:

\begin{restatable}{lemma}{lemOne}
Suppose $T : \mathbb{R}^p \to \mathbb{R}^m$ is a measurable and injective function. Then it always holds that
\[
P(Y \mid X) = P(Y \mid Z) \quad \text{almost surely},
\]
and in particular $Z$ is sufficient for $Y$.
\end{restatable}
Moreover, any compositions of injective transformations remains injective and is also sufficient. 

\subsection{Classical Neural Network Architecture}

Classical neural networks usually just include fully connected layers, activation functions, and possibly softmax layer at the end for classification tasks. For the purpose of subsequent theorems, we opt to use random variables (as row-vector for multivariate variable) throughout this paper. 

\paragraph*{Fully Connected Layers}
A fully connected layer applies an affine transformation to the input. Given input variable $X \in \mathbb{R}^{1 \times p}$, a fully connected layer (without activation function) can be defined as
\[
Z = X \alpha + \beta,
\]
where $\alpha \in \mathbb{R}^{p \times m}$ is a weight matrix, and $\beta \in \mathbb{R}^{1 \times m}$ is a row vector. The output $Z \in \mathbb{R}^{m}$ is a random variable of dimension $1 \times m$. This single layer is parameterized by $\alpha$ and $\beta$, which are learned during training to minimize a loss function. 

\paragraph*{Activation Functions}
Activation function is usually applied to the output of a layer elementwise, which introduce nonlinearity and allow neural networks to approximate complex functions. The ReLU function $\sigma(z) = \max(0, z)$, is the most common in modern architecture due to its simplicity, which is non-invertible and zeroes out all negative values. The identity activation $\sigma(z) = z$ leaves the input unchanged. The sigmoid function $\sigma(z) = \frac{1}{1 + e^{-z}}$, is a traditional choice, which transforms each coordinate into $(0,1)$ and is strictly monotonic and continuous. 

\paragraph*{Multiple Layers and Composition}
Classical neural networks are mostly formed by composing multiple fully connected layers, typically alternating between single layers and nonlinear activations:
\[
Z^{(1)} = \sigma^{(1)}(X \alpha^{(1)} + \beta^{(1)}), \quad \dots, \quad Z^{(L)} = \sigma^{(L)}(Z^{(L-1)} \alpha^{(L)} + \beta^{(L)}).
\]
Here,
\begin{itemize}
    \item $X \in \mathbb{R}^{1 \times p}$ is the input random variable
    \item $\alpha^{(\ell)} \in \mathbb{R}^{m^{(\ell-1)}*m^{(\ell)}}$ and $\beta^{(\ell)} \in \mathbb{R}^{1*m^{(\ell)}}$ are weights and biases at layer $\ell$, where $m^{(\ell)}$ is the neuron size at layer $\ell$ and $m^{0}=p$.
    \item $\sigma^{(\ell)}$ is an elementwise activation function (e.g., ReLU, sigmoid).
\end{itemize}

\paragraph*{Learning and Training Objective}
All trainable parameters in a neural network, including weights $\{\alpha^{(\ell)}\}$, biases $\{\beta^{(\ell)}\}$, are learned from sample data. Training is typically performed via stochastic gradient descent or its variants, optimizing a task-specific loss function, usually either the squared error loss in regression, or cross-entropy loss in classification, as well as custom losses such as contrastive loss, KL divergence, etc.,  depending on the problem setting. The optimization algorithm randomly initialize initial parameters, forward the training data into the model, then back-propagate to update parameters by the previous estimator minus learning rate times derivatives. The final model is determined upon some stopping criterion is reached.

\section{Graph Variable and Anchor Points}
\label{sec:graphassumption}

In this section, we formalize a graph variable from classical neural networks. This graph perspective allows us to view each layer in fully connected neural network as producing a new random variable that is graph-like, based on the input variable and a set of anchor points, setting the foundation for achieving sufficiency via either injectivity and region separation in later sections.

\subsection{Graph Perspective of Single Layer}
We will start with a single fully connected layer. A fully connected layer with activation is written as:
\[
Z = \sigma(X \alpha + \beta) \in \mathbb{R}^{m}.
\]
Equivalently, we can interpret this layer as constructing a graph variable $Z$, via the random variable $X$, the anchor points $\alpha_{j}$ (each column of $\alpha$) for $j=1,\ldots,m$, and a pairwise graph function $A(x, \alpha) = \sigma(\langle x, \alpha \rangle + \beta)$, such that
\[
Z = [A(X,\alpha_1), \dots, A(X,\alpha_m)] \in \mathbb{R}^{m}. 
\]
This can be viewed as first forming an inner product similarity between the input and all anchor points, then threshold-ed by the term $\beta$ to adjusts the edge weight between $X$ and $\alpha$, and the ReLU function ensures non-negative edge weights. The output $Z \in \mathbb{R}^{m}$, therefore, can be viewed as a graph variable that is aggregated, adjusted, with ensured non-negativity entry-wise. 

While the above formulation very well suits a high-level analysis, a more granular view can be more convenient at times. We may assume $A(x, \alpha) = \langle x, \alpha \rangle$, form the graph variable $Z$ as concatenations of $\{A(X,\alpha_j)\}$, then let $\tilde{Z}=\sigma(Z + \beta)$. This view better connects with standard pairwise function, where $A(\cdot,\cdot): \mathbb{R}^p \times \mathbb{R}^p \rightarrow \mathbb{R}$ can simply assume classical pairwise functions, some standard examples include:
\begin{itemize}
    \item Inner product: $A(x, \alpha) = \langle x, \alpha \rangle$
    \item Euclidean distance (squared): $A(x, \alpha) = \|x - \alpha\|^2$
    \item Cosine similarity: $A(x, \alpha) = \frac{\langle x, \alpha \rangle}{\|x\| \|\alpha\|}$
    \item Gaussian kernel: $A(x, \alpha) = \exp\left(-\frac{\|x - \alpha\|^2}{2\sigma^2}\right)$
\end{itemize}
In general, one can use any distance metric, kernels, dissimilarity functions here. Inner product is the dominating choice in classical neural network, because of the convenience in gradient calculation. 

We shall use the granular view, i.e., separately analyze inner product, intercept, activation, in Section~\ref{sec:singlelayer} for single-layer analysis; and consider the more general graph view in Section~\ref{sec:multilayer} for multi-layer analysis.

\subsection{Dense Support for Initialized Anchor Points}

Next, we consider the property of each anchor points $\{\alpha_{j}, j=1,\ldots,m\}$, which are columns of $\alpha$. In practice, those anchor points are learned parameters from training data. Therefore, to ensure neural network models can preserve sufficiency, we need to investigate proper conditions on $\{\alpha_{j}\}$. 

A key result needed for Section~\ref{sec:singlelayer} and Section~\ref{sec:multilayer} is that the anchor points, if we let $m$ goes to infinite,  are densely supported in the support of $X$ --- or more generally, the anchor points $\alpha^{(\ell)}$ of current layer are densely supported in the support of prior input $Z^{(\ell-1)}$. This can be easily proved if the anchor points are independently and identically distributed:

\begin{restatable}{lemma}{lemTwo}
\label{cor:suff-iid-fx}
Suppose anchor points $\{\alpha_j\}_{j=1}^m$ are drawn i.i.d. from the same distribution as $X$, with distribution $F_X$ supported on a compact subset of $\mathbb{R}^p$. As $m \to \infty$, the set $\{\alpha_j\}_{j=1}^m$ becomes dense in the support of $X$ with probability approaching 1.
\end{restatable}

The iid assumption of $\alpha_j$, and it sampled from $F_X$, matches with parameter initialization in neural network \cite{rahimi2007random,jacot2018ntk}, and also in high-capacity overparameterized regimes \cite{allen2019convergence,belkin2019reconciling}. 

\subsection{Dense Support for Learned Anchor Points}

We would like the same dense support to always hold for the anchor points in subsequent model training and backpropagation. To justify that, let $\mathcal{L}(Z, Y)$ denote the loss function, the update rule under gradient descent is:
\[
\alpha_j^{t+1} = \alpha_j^{t} - \eta \cdot \frac{\partial \mathcal{L}}{\partial \alpha_j},
\]
with $\eta > 0$ the learning rate. Note that for this subsection only, the superscripts $\cdot^{t+1}$ denotes the current training iteration, not to be confused with layer notation. Clearly, this update depends on the current data batch $X$, labels $Y$, and the anchor vector $\alpha_j$ itself, but not directly on any other $\alpha_k$ for $k \ne j$, except through possibly empirical loss from previous iteration. The next lemma proves when the back propagation preserves dense support of anchor points.

\begin{restatable}{lemma}{lemThree}
\label{lem:density-grad}
Let $\{\alpha_j^0\}_{j=1}^m$ be drawn i.i.d. from $F_X$, the distribution of $X$ with compact support $\operatorname{supp}(F_X) = \operatorname{supp}(X) \subset \mathbb{R}^p$. Define the gradient update:
\[
\alpha_j^{t+1} = \alpha_j^t - \eta \cdot \frac{\partial \mathcal{L}}{\partial \alpha_j}, \quad j = 1, \dots, m,
\]
for a differentiable loss $\mathcal{L} = \mathcal{L}(X, Y; \{\alpha_j\}, \theta)$.

Suppose:
\begin{itemize}
    \item[(i)] $\mathcal{L}$ is differentiable in each $\alpha_j$ and jointly continuous in $(X, \{\alpha_j\}, \theta)$;
    \item[(ii)] For each $t$, the iterates $\{\alpha_j^t\}$ remain in a compact superset $\operatorname{supp}(X)^t \supseteq \operatorname{supp}(X)$;
    \item[(iii)] Each update map $f_t: \alpha \mapsto \alpha - \eta \cdot \partial \mathcal{L}/\partial \alpha$ is a homeomorphism from $\operatorname{supp}(X)^t$ onto its image.
\end{itemize}

Then, as $m \to \infty$, with probability 1, the sets $\{\alpha_j^t\}_{j=1}^m$ are dense in $\operatorname{supp}(X)^t$ for all $t \geq 0$.
\end{restatable}

Assumption (i) of Lemma~\ref{lem:density-grad} — that the loss $\mathcal{L}$ is differentiable in each anchor point $\alpha_j$ and continuous in all inputs — is satisfied for standard smooth loss functions such as mean squared error and cross-entropy, and for neural network outputs composed of linear layers, affine transforms, and any smooth activation functions.

Assumption (ii), which requires that $\{\alpha_j^{t}\}$ remain within a compact superset of the data support $\operatorname{supp}(X)$, is typically satisfied in practice under bounded weight initialization and common training conditions. In particular, inputs $X$ are supported on a compact set, parameters are initialized from compactly supported distributions (e.g., uniform or truncated Gaussian), and learning rate and gradient clipping prevent parameter explosion, ensuring that iterates remain within a bounded region.

Assumption (iii) asserts that the gradient-based update map does not collapse the anchor points, i.e., the mapping remains injective and continuous, thereby preserving the topological structure (and density) of the initial anchor configuration. While global injectivity is difficult to guarantee in full generality, it is reasonable to assume local injectivity and continuity of the gradient update in common training regimes, such as in the lazy training regime \cite{chizat2019lazy} and the neural tangent kernel regime \cite{jacot2018ntk,lee2019wide}.

These perspectives suggest that for appropriately tuned learning rates and smooth loss landscapes, the gradient flow behaves as a locally injective and continuous transformation, and therefore preserves the denseness of anchor points in $\operatorname{supp}(X)$ throughout training.

\section{Asymptotic Sufficiency for Fully Connected Hidden Layer}
\label{sec:singlelayer}

In this section we present a sequence of results for asymptotic sufficiency of single-layer neural representations. Based on the graph variable perspective and anchor points being dense from Section~\ref{sec:graphassumption}, we begin with asymptotic sufficiency for general pairwise function, then extend to specific corollaries for inner product, distance-based, and kernel-based transformations, then include activation functions such as ReLU or sigmoid, and a combination of the above to characterize sufficiency for $Z = \sigma(X \alpha + \beta)$.

\subsection{Injectivity-Based Sufficiency}
\begin{restatable}{theorem}{thmOne}
\label{thm:asymp-suff-dense}
Let $X \in \mathbb{R}^p$ be a random variable supported on a compact set $\operatorname{supp}(X)$, and let $Y$ be any target variable of finite moments. Let $\{\alpha_j\}_{j=1}^\infty \subset \mathbb{R}^p$ be dense in $\operatorname{supp}(X)$, and let $A : \mathbb{R}^p \times \mathbb{R}^p \to \mathbb{R}$ be a measurable function satisfying:

\begin{itemize}
    \item[(i)] For any $x_1 \ne x_2$ in $\operatorname{supp}(X)$, there exists $\alpha \in \operatorname{supp}(X)$ such that $A(x_1, \alpha) \ne A(x_2, \alpha)$;
    \item[(ii)] For each $x \in \operatorname{supp}(X)$, the map $\alpha \mapsto A(x, \alpha)$ is continuous;
    \item[(iii)] Each $A(X, \alpha_j)$ is measurable and has finite second moment.
\end{itemize}

Define $Z_m := [A(X, \alpha_1), \dots, A(X, \alpha_m)] \in \mathbb{R}^m$. Then, as $m \to \infty$, $Z_m$ becomes injective on $\operatorname{supp}(X)$, and
\[
\lim_{m \to \infty} \left\| \mathbb{P}(Y \mid Z_m) - \mathbb{P}(Y \mid X) \right\|_{L^2(F_X)} = 0.
\]
\end{restatable}

\subsection{Asymptotic Sufficiency under Specific Pairwise Functions}

Theorem~\ref{thm:asymp-suff-dense} is in fact a general result for any function $A(\cdot,\cdot)$. Namely, if the pairwise similarity satisfies the required condition, then the resulting graph variable $Z$ is asymptotically sufficient for infinite $m$. The next corollary verifies that standard pairwise functions, from inner product to distance to universal kernel, satisfy the required conditions.

\begin{restatable}{theorem}{thmTwo}
\label{cor:sufficiency-common}
Let $X \in \mathbb{R}^p$ be a random variable supported on a compact set $\operatorname{supp}(X) \subset \mathbb{R}^p$, and let $\{\alpha_j\}_{j=1}^\infty \subset \mathbb{R}^p$ be a sequence of anchor points dense in $\operatorname{supp}(X)$. Then the representation
\[
Z_m = \left[ A(X, \alpha_1), \dots, A(X, \alpha_m) \right] \in \mathbb{R}^m
\]
is asymptotically sufficient for $X$ in each of the following cases:
\begin{itemize}
    \item[(a)] $A(X, \alpha) = \langle X, \alpha \rangle$,
    \item[(b)] $A(X, \alpha) = \|X - \alpha\|$,
    \item[(c)] $A(X, \alpha) = \|X - \alpha\|^2$,
    \item[(d)] $A(X, \alpha) = k(X, \alpha)$, where $k$ is a continuous universal kernel on $\operatorname{supp}(X)$,
    \item[(e)] $A(X, \alpha) = \frac{\langle X, \alpha \rangle}{\|X\| \|\alpha\|}$, under the assumption that $Y \perp \|X\| \mid \tilde{X}$.
\end{itemize}
Moreover, in case (e), if $\|X\|$ carries information about $Y$, then the augmented representation $[Z_m; \|X\|]$ is asymptotically sufficient.
\end{restatable}

Next, a fully connected layer usually comes with an intercept term, which is a learned parameter but still preserves sufficiency. 

\begin{restatable}{corollary}{corOne}
\label{cor:suff-intercept}
Following the same conditions and notations as in Theorem~\ref{thm:asymp-suff-dense}, further define:
\[
\tilde{Z} = Z + \beta, \quad \beta \in \mathbb{R}^{m}.
\]
Then, as $m \to \infty$, the map $X \mapsto \tilde{Z}$ becomes injective on the support of $X$:
\[
\lim_{m \to \infty} \| P(Y \mid \tilde{Z}) - P(Y \mid X) \|_{L^2(F_X)} = 0.
\]
\end{restatable}

\subsection{Asymptotic Sufficiency under Nonlinear Activations}

Next, we aim to extend the injectivity-based asymptotic sufficiency to include nonlinear activation function. If using identity activation function, and there is nothing to prove. The sigmoid function is strictly increasing, continuous, and bijective from $\mathbb{R} \to (0,1)$ --- which means if $Z$ is sufficient for $X$, $\sigma(Z)$ is also sufficient for $X$. Therefore, all we need is to justify the ReLU activation function, which is not injective in general, as stated in the following lemma:

\begin{restatable}{lemma}{lemFour}
Let $Z$ be sufficient for $X$ with respect to $Y$.

\begin{itemize}
    \item[(a)] If $\sigma : \mathbb{R} \to \mathbb{R}$ is bijective and measurable (e.g., identity, sigmoid), then $\sigma(Z)$ is always sufficient.
    \item[(b)] If $\sigma$ is not injective (e.g., ReLU), then sufficiency may not be preserved unless $Z$ lies in a subset of $\mathbb{R}^m$ on which $\sigma$ is injective.
\end{itemize}
\end{restatable}

However, even with information-destroying $\sigma$ (like ReLU), if we use enough such features, we can still recover all the information about $X$ asymptotically, as proved in the next theorem: 

\begin{restatable}{theorem}{thmThree}
\label{thm:relu-injective}
Under the same assumptions as Theorem~\ref{thm:asymp-suff-dense}, define the ReLU-activated representation:
\[
\tilde{Z} = \sigma(Z + \beta), \quad \text{where } Z = [A(X, \alpha_1), \dots, A(X, \alpha_m)] \in \mathbb{R}^m, \text{ and } \beta \in \mathbb{R}^m.
\]
Here, $\sigma$ denotes the ReLU function $\sigma(u) = \max(0, u)$ applied elementwise. Then, as $m \to \infty$, the map $X \mapsto \tilde{Z}$ becomes injective on the support of $X$, and:
\[
\lim_{m \to \infty} \left\| \mathbb{P}(Y \mid \tilde{Z}) - \mathbb{P}(Y \mid X) \right\|_{L^2(F_X)} = 0.
\]
\end{restatable}

\section{Classical Multi-Layer Neural Networks and Asymptotic Sufficiency}
\label{sec:multilayer}

We now extend our sufficiency results to classical multi-layer neural networks.

\subsection{Graph Perspective of Multi-Layer Representations}

The classical multi-layer network typically consists of multiple fully connected layers and activation function for each. This structure means $\ell=1,\ldots,L$ layers, let $Z^{(0)} := X \in \mathbb{R}^p$ and recursively define:
\[
Z^{(\ell)} = \sigma^{(\ell)} \left( Z^{(\ell-1)} \alpha^{(\ell)} + \beta^{(\ell)} \right),
\]
where:
\begin{itemize}
  \item $\alpha^{(\ell)} \in \mathbb{R}^{m^{(\ell-1)} \times m^\ell}$ is a matrix of anchor columns,
  \item $\beta^{(\ell)} \in \mathbb{R}^{1 \times m^\ell}$ is a bias vector,
  \item $\sigma^{(\ell)}$ is an elementwise activation function (e.g., ReLU, sigmoid),
  \item $m^0 = p$, $m^\ell$ is the width of layer $\ell$.
\end{itemize}

We already tackled each component, from affine transformation to possibly other pairwise similarity function, as well as adding intercept, and activation function, in Section~\ref{sec:singlelayer}, which allows us to adopt a more general graph view in this section. In a multi-layer architecture, we can view each $Z^{(\ell)}$ as a graph variable, based on the previous graph variable $Z^{(\ell-1)}$ and a new set of anchors $\alpha^{(\ell)}$ drawn from $F_{\alpha^{(\ell)}} \in \mathbb{R}^{m^{(\ell-1)}}$. Namely, 
\[
Z^{(\ell)} = [A^{(\ell)}(Z^{(\ell-1)},\alpha_1^{(\ell)}), \dots, A^{(\ell)}(Z^{(\ell)},\alpha_{m^{(\ell)}}^{(\ell)})] \in \mathbb{R}^{m^{(\ell)}}. 
\]
Here, the pairwise function $A^{(\ell)}$ can be different in each layer. This recursive graph embedding induces a nested graph structure over layers, where each new representation can be seen as a graph transformation over a previous graph variable. 

\subsection{Layerwise Preservation of Sufficiency}

We now show that if each layer satisfies the conditions from the single layer case, i.e., point-separating feature maps, coupled with proved dense support in Lemma~\ref{lem:density-grad}, the resulting multi-layer neural network is asymptotically sufficient for any finite number of layers.

\begin{restatable}{theorem}{thmFour}
\label{thm:multi-layer-general}
Let $X \in \mathbb{R}^p$ be a random variable supported on a compact set $\operatorname{supp}(X) \subset \mathbb{R}^p$, and let $Y$ be any target variable. Define a sequence of layerwise representations:
\[
Z^{(0)} := X, \quad Z^{(\ell)} := \left[ A^{(\ell)}\left(Z^{(\ell-1)}, \alpha_1^{(\ell)}\right), \dots, A^{(\ell)}\left(Z^{(\ell-1)}, \alpha_{m^{(\ell)}}^{(\ell)}\right) \right] \in \mathbb{R}^{m^{(\ell)}}, \quad \ell = 1, \dots, L,
\]
where $L$ is a fixed positive integer, each $A^{(\ell)} : \mathbb{R}^{m^{(\ell-1)}} \times \mathbb{R}^{m^{(\ell-1)}} \to \mathbb{R}$ defines a neuron-wise feature map, and each $\alpha_j^{(\ell)}$ is a learned parameter (e.g., weight or embedding vector) such that:

\begin{itemize}
    \item[(i)] $A^{(\ell)}(z, \alpha)$ is measurable in $(z, \alpha)$ and continuous in $\alpha$ for each fixed $z$;
    \item[(ii)] For any distinct $z_1, z_2$ in the support of $Z^{(\ell-1)}$, there exists $\alpha$ such that $A^{(\ell)}(z_1, \alpha) \ne A^{(\ell)}(z_2, \alpha)$;
    \item[(iii)] The anchor parameters $\{\alpha_j^{(\ell)}\}_{j=1}^\infty$ remain dense in $\operatorname{supp}(Z^{(\ell-1)})$ with probability 1 under initialization and gradient updates, as in Lemma~\ref{lem:density-grad}.
\end{itemize}

Then, with probability 1, the final-layer representation $Z^{(L)}$ is asymptotically sufficient for $X$ as all layer widths grow:
\[
\lim_{m^{(1)}, \dots, m^{(L)} \to \infty} \left\| \mathbb{P}(Y \mid Z^{(L)}) - \mathbb{P}(Y \mid X) \right\|_{L^2(F_X)} = 0.
\]
\end{restatable}
This result applies to any composition of single-layer components previously analyzed — including inner products, additive intercepts, ReLU activations, or kernel evaluations — as long as each neuron’s feature map $A^{(\ell)}$ satisfies the continuity and separation conditions. Furthermore, since each $\alpha_j^{(\ell)}$ can correspond to a neuron’s parameters at any stage of training, this theorem implies that information about $X$ is preserved across layers and across training iterations, from initialization to convergence. Note that the finiteness of $L$ is essential here to ensure convergence, as it allows the sum of layerwise errors to vanish. 

\section{Exact Sufficiency with Region Separation}
\label{sec:region}

The injectivity-based sufficiency thus far requires the pairwise function to be point-separating, and the number of anchor points increases to infinity so it is dense in the support of $X$. However, in real-world neural network architectures, the number of neurons per layer is always finite. 

In order to better align the theoretical results with practice, we start by considering an important special case: when the input random variable $X$ takes on only $m$ discrete values, one can use $m$ anchor points from each value of $X$, and sufficiency can be achieved exactly using $m$ neurons. 

This motivates our next investigation of sufficiency through region-separating representations: instead of requiring infinite anchor points over the full support of $X$, we aim to show that neural networks can distinguish the relevant statistical modes of $X$ that determine $P(Y \mid X)$ --- which means neural network does not need to be injective, but still preserves sufficiency (or approximately so) for structured random variables. 

We shall first consider general transformation $T$, not just neural network, and pinpoint the region-separating property to ensure sufficiency. Then we prove that fully-connected layer and convolutional layer are region-separating transformation in the next two sections.

\subsection{Exact Sufficiency with Discrete Support}

\begin{restatable}{lemma}{lemFive}
\label{lem:discrete-support}
Let $X$ be a random variable taking values in a finite set $\{x_{1}, \dots, x_{m}\} \subset \mathbb{R}^p$, and let $Y$ be any target variable. Suppose $T : \mathbb{R}^p \to \mathbb{R}^d$ is a measurable function such that:
\[
T(x_{i}) \ne T(x_{j}) \quad \text{for all } i \ne j.
\]
Then the representation $Z = T(X)$ is sufficient for $X$ with respect to $Y$, i.e.,
\[
P(Y \mid Z) = P(Y \mid X).
\]
\end{restatable}

When $X$ is discrete, the setting of Lemma~\ref{lem:discrete-support} is simple enough to allow for sufficiency with finite neuron count in a single layer. For example. let $T(X) = [\langle X, x_{1} \rangle, \dots, \langle X, x_{m} \rangle]$. Since the vectors $\{x_{i}\}$ are distinct, the resulting vector $T(X)$ is injective on $\{x_{1}, \dots, x_{m}\}$ as long as the Gram matrix is nonsingular (which holds if the $x_{i}$ are linearly independent or in general position).

\subsection{Defining Region Separation}

Next, we extend the structured variable $X$ to be more realistic --- instead of being discrete, they can be continuous, but its support can be partitioned into $m$ disjoint regions where each region has constant conditional distribution. This arises naturally, for example, in piecewise-constant classification, such as decision trees where the model attempt to split the support into pure leaf region; and also in piecewise-linear regression, where $f(X)$ is linear or constant within each region. To that end, we define what we meant for region-separated random variable and region-separating transformation.

\begin{definition}[Region-Separated Random Variable]
Given a random variable $X$ in $\mathbb{R}^p$, whose support can be partitioned into $m$ compact and measurable regions $\{\mathcal{R}_1, \dots, \mathcal{R}_m\}$ such that for each $i$, the conditional distribution $P(Y \mid X = x)$ is constant over $\mathcal{R}_i$:
\[
\forall x \in \mathcal{R}_i, \quad P(Y \mid X = x) = q_i.
\]
Furthermore, the regions $\{\mathcal{R}_1, \dots, \mathcal{R}_m\}$ have non-empty interior and their intersections are of zero-measure, i.e.,
\[
\text{For all } i \ne j, \quad Prob(X \in \mathcal{R}_i \cap \mathcal{R}_j) = 0.
\]
Then we say $X$ is a region-separated random variable with $m$ regions. 
\end{definition}

\begin{definition}[Region-Separating Transformation]
Given a region-separated random variable $X$ with $m$ regions $\{\mathcal{R}_1, \dots, \mathcal{R}_m\}$, we say $T: \mathbb{R}^p \to \mathbb{R}^d$ is a region-separating transformation for $X$, if and only if
\[
\forall i \ne j, \forall x \in \mathcal{R}_i, x' \in \mathcal{R}_j, \quad T(x) \ne T(x').
\]
\end{definition}

\begin{restatable}{theorem}{thmFive}
\label{lem:region-constant}
Given a region-separated random variable $X$ with $m$ regions. If the transformation $T$ is region-separating for $X$, then $Z = T(X)$ is sufficient for $X$ with respect to $Y$.
\end{restatable}
The requirement that $T$ is region-separating is notably weaker than requiring $T$ to be injective on the support of $X$. Injectivity demands that $T(x) \neq T(x')$ for all $x \neq x'$, but region separation only asks that $T(x)$ differs whenever $x$ belongs to a different region $\mathcal{R}_i$ that determines $P(Y \mid X)$. Within each region, $T$ is free to collapse — for example, mapping all of $\mathcal{R}_i$ to a single position $z_i$. 

This is somewhat related the recently observed phenomenon of neural collapse in deep classification networks \cite{papyan2020neural, han2021neural}: It has been empirically observed that representations of inputs from the same class collapse to a single prototype vector, while inter-class prototypes become maximally separated. A region-separating transformation does support this empirical observation: collapsed representation can be sufficient and preserves all information for prediction, in case of region-separated random variables, which is supported by Figure~\ref{fig1} in simulation.


\subsection{Approximated Region Separation}

Instead of requiring the conditional distribution to be identical within each region, we may allow for bounded variation within each region.

\begin{definition}
Given a random variable $X$ in $\mathbb{R}^p$, whose support can be partitioned into $m$ compact and measurable regions $\{\mathcal{R}_1, \dots, \mathcal{R}_m\}$ such that for each $i$, the conditional distribution $P(Y \mid X = x)$ is constant over $\mathcal{R}_i$ up-to $\epsilon>0$:
\[
\forall x, x' \in \mathcal{R}_i, \quad \|P(Y \mid x) - P(Y \mid x')\|_1 \leq \epsilon.
\]
Furthermore, the regions $\{\mathcal{R}_1, \dots, \mathcal{R}_m\}$ have non-empty interior and their intersections are of zero-measure, i.e.,
\[
\text{For all } i \ne j, \quad Prob(X \in \mathcal{R}_i \cap \mathcal{R}_j) = 0.
\]
Then we say $X$ is an $\epsilon$-region-separated random variable with $m$ regions. 
\end{definition}

With mild condition, any random variable $X$ can be viewed as $\epsilon$-region-separated, but possibly with infinite $m$.

\begin{restatable}{theorem}{thmSix}
\label{thm:universal-eps-region}
Let $X \in \mathbb{R}^p$ be a random variable supported on a compact set $\mathcal{X} \subset \mathbb{R}^p$, and let $f(x) := \mathbb{P}(Y \mid X = x)$ denote the conditional distribution of $Y$ given $X$. Suppose $f \in L^1(F_X)$, i.e., $f$ is integrable with respect to the marginal distribution of $X$.

Then for any $\epsilon > 0$, there exists a finite measurable partition $\{\mathcal{R}_1, \dots, \mathcal{R}_m\}$ of $\mathcal{X}$ such that:
\[
\forall i \in \{1, \dots, m\}, \quad \forall x, x' \in \mathcal{R}_i, \quad \|\mathbb{P}(Y \mid x) - \mathbb{P}(Y \mid x')\|_1 \leq \epsilon.
\]
That is, $X$ is $\epsilon$-region-separated with respect to the partition $\{\mathcal{R}_i\}_{i=1}^m$.
\end{restatable}

\subsection{Probability Bound under Approximated region separation}
Next, we prove a general sufficiency result regarding an $\epsilon$-region-separated random variable $X$. For any region-separating transformation $T$ (with respect to the regions of $X$), it can preserve the conditional distribution up-to an error bound that is controlled by the maximum deviation $\epsilon$ of the conditional distributions within each region. 

\begin{restatable}{theorem}{thmSeven}
\label{thm:eps-region-preserved}
Let $X$ be an $\epsilon$-region-separated random variable with $m$ regions $\{\mathcal{R}_1, \dots, \mathcal{R}_m\}$. Suppose $T(X)$ is a region-separating transformation; that is,
\[
\forall i = 1, \dots, m, \quad \forall x, x' \in \mathcal{R}_i, \quad T(x) = T(x') := z_i.
\]

Then the transformed variable $Z := T(X)$ is also $\epsilon$-region-separated. Specifically:
\begin{itemize}
    \item The support of $Z$ is finite and contained in $\{z_1, \dots, z_m\}$.
    \item For all $i$ and all $x \in \mathcal{R}_i$, we have:
    \[
    \| \mathbb{P}(Y \mid Z = z_i) - \mathbb{P}(Y \mid x) \|_1 \leq \epsilon.
    \]
\end{itemize}
\end{restatable}

Therefore, when the random variable $X$ is region-separated, and the transformation $T(\cdot)$ is region-separating for $X$, then, $T(X)$ would remain region-separated. This is true for the general $\epsilon$-region-separated variable. 

If we think about this result in terms of neural network layers, if every single layer ends up being region-separating, then the output of each layer remains region-separating and preserve sufficiency up-to the error $\epsilon$.

\section{Sufficiency and Optimal Loss Bound of Neural Networks}
\label{sec:regionLayer}

\subsection{Sufficiency for Fully Connected Layer}

Based on the results for general transformation, it suffices to prove that a single full connected layer is region-separating, which then qualify it, as well as any multi-layer composition, as preserving conditional density for region-separated random variables and therefore sufficient. From here onwards, we further require that each region be convex, as this is necessary to enable explicit construction of anchor points in neural networks.

\begin{restatable}{theorem}{thmEight}
\label{thm:linear-region-separation-refined}
Let $X \in \mathbb{R}^p$ be an $\epsilon$-region-separated random variable with $m$ measurable regions $\{\mathcal{R}_1, \dots, \mathcal{R}_m\}$, each is convex and of positive measure.

Then there exists a linear map $Z(x) = [\langle x, \alpha_1 \rangle + b_1, \dots, \langle x, \alpha_m \rangle + b_m] \in \mathbb{R}^m$ such that:
\[
x \in \mathcal{R}_i,\ x' \in \mathcal{R}_j,\ i \ne j \quad \Rightarrow \quad Z(x) \ne Z(x') \quad \text{almost surely}.
\]
That is, $Z$ separates the regions almost everywhere.
\end{restatable}

\begin{restatable}{theorem}{thmNine}
\label{thm:relu-region-separation}
Let $X \in \mathbb{R}^p$ be an $\epsilon$-region-separated random variable with $m$ measurable regions $\{\mathcal{R}_1, \dots, \mathcal{R}_m\}$, each is convex and of positive measure.

Then there exist vectors $\{\alpha_j\}_{j=1}^m \subset \mathbb{R}^p$ and biases $\{b_j\}_{j=1}^m \subset \mathbb{R}$ such that the ReLU-activated representation
\[
Z(x) := \left[ \max(0, \langle x, \alpha_1 \rangle + b_1), \dots, \max(0, \langle x, \alpha_m \rangle + b_m) \right] \in \mathbb{R}_{\ge 0}^m
\]
is region-separating almost surely. That is,
\[
x \in \mathcal{R}_i,\ x' \in \mathcal{R}_j,\ i \ne j \quad \Rightarrow \quad Z(x) \ne Z(x') \quad \text{almost surely}.
\]
\end{restatable}
Now, combined with Theorem~\ref{thm:eps-region-preserved}, we can readily conclude that neural networks remain sufficient, or approximately sufficient, even as the number of layers increases, potentially to infinity. This is a desirable property that aligns well with modern deep architectures.
 
\subsection{Optimal Loss Bound on Regression and Classification}

Theorem~\ref{thm:universal-eps-region} showed that any random variable can be represented as $\epsilon$-region-separated. Combined with Theorems~\ref{thm:eps-region-preserved} and~\ref{thm:relu-region-separation}, we immediately obtain the following corollary: a neural network is a sufficient transformation when $\epsilon=0$, or approximately so with at most $\epsilon$ deviation in conditional probability.

\begin{restatable}{corollary}{corTwelve}
\label{cor:nn-eps-sufficiency}
Let $X$ be an $\epsilon$-region-separated random variable with $m$ convex regions $\{\mathcal{R}_1, \dots, \mathcal{R}_m\}$, and $T(X)$ denotes a ReLU neural network with at least $m$ neurons in each hidden layer, where all parameters satisfy Theorem~\ref{thm:relu-region-separation} to ensure region-separating.

Then the output $Z = T(X)$ is sufficient for $X$ when $\epsilon = 0$; and in general approximately sufficient with bound:
\[
\| P(Y \mid Z) - P(Y \mid X) \|_1 \leq \epsilon.
\]
\end{restatable}

Next, we relate the probability bound to the Bayes-optimal loss in both regression and classification settings. Background on optimal loss, i.e., mean squared error for regression and Bayes classification error under random variables, can be found in \cite{hastie2009elements, DevroyeGyorfiLugosiBook}.

\begin{restatable}{theorem}{thmTwelve}
\label{thm:error-bound-reg-class}
Let $X$ be an $\epsilon$-region-separated random variable with $m$ convex regions $\{\mathcal{R}_1, \dots, \mathcal{R}_m\}$, and let $T(X)$ denote a ReLU neural network with at least $m$ neurons in each hidden layer, where all parameters satisfy the conditions of Theorem~\ref{thm:relu-region-separation} to ensure region separation. Define $Z := T(X)$.

Then the following bounds on prediction loss hold:

\textbf{(1) Regression:} If $Y \in [y_1, y_2]$ and a regression neural network is trained, then the optimal mean squared error satisfies
\[
\left| \operatorname{MSE}^*(Z) - \operatorname{MSE}^*(X) \right| \leq \epsilon^2 (y_2-y_1)^2 + \epsilon(y_2 - y_1)^2/2,
\]
where $\operatorname{MSE}^*(X) := \mathbb{E}[(Y - \mathbb{E}[Y \mid X])^2]$.

\textbf{(2) Classification:} If $Y \in \{1, \dots, K\}$ and a classification neural network is trained, then the Bayes-optimal classification error satisfies
\[
\left| \operatorname{Err}^*(Z) - \operatorname{Err}^*(X) \right| \leq \epsilon,
\]
where $\operatorname{Err}^*(X) := 1 - \mathbb{E}_X\left[ \max_y P(Y = y \mid X) \right]$.
\end{restatable}

In the special case where $\epsilon = 0$, the neural network is exactly sufficient, and the Bayes-optimal loss remains unchanged. The theorem also showed that in classification problems, the optimal loss is robust against small deviations. For example, a region-separation imperfection of $\epsilon = 0.001$ leads to at most a $0.001$ increase in Bayes-optimal classification error. However, in regression tasks, the same imperfection can lead to a larger increase in optimal mean squared error, as the error scales with the range of $Y$. This may partially explain why neural networks tend to achieve more success and breakthroughs in classification-type tasks, particularly those involving categorical outputs, compared to regression tasks.

\subsection{On the Number of Anchor Points}

In our explicit constructions, we show that if a random variable $X$ is $\epsilon$-region-separated with $m$ measurable regions $\{\mathcal{R}_1, \dots, \mathcal{R}_m\}$, then a representation map using $m$ neurons (or anchor points) is sufficient to asymptotically preserve region separation. Each neuron is explicitly designed to distinguish one region from the others, leading to a region-separating map $Z(X)$ where each region is mapped to a distinct code. 

When $X$ is $m$-region-separated, it is possible to use fewer than $m$ neurons. For instance, if the regions are geometrically aligned or linearly separable in groups, a smaller set of neurons can distinguish them by composing simple linear cuts. 
In practice, however, the number of regions $m$ is unknown, and the true data distribution may contain finer substructure or continuous variation that is not captured by a fixed partition. For this reason, it is common to begin with a generously sized initial layer, often with width far exceeding the number of output classes, enabling the network to detect latent subregions and preserve flexibility for downstream compression \cite{neyshabur2018role}. This practice reflects both theoretical insights and empirical heuristics in modern neural networks.

\section{Graph Sufficiency for Convolution}
\label{sec:conv}
In this section, we extend the region separation based sufficiency into convolutional neural networks (CNNs), a layer design working particularly well for structured inputs such as images.
We first reinterpret convolutional layer through our anchor-point framework between input random variable and sparse anchor points. 

In CNN, inputs typically take the form of 3D tensors $X \in \mathbb{R}^{H \times W \times C}$, representing height, width, and channels. A 2D convolutional layer applies learned filters (kernels) $K \in \mathbb{R}^{k_H \times k_W \times C}$ over sliding spatial windows of $X$, producing output feature maps.

Formally, for stride $s$ and padding $p$, the output spatial size is:
\[
\hat{H} = \left\lfloor \frac{H - k_H + 2p}{s} \right\rfloor + 1, \quad
\hat{W} = \left\lfloor \frac{W - k_W + 2p}{s} \right\rfloor + 1.
\]

Each output activation is computed by sliding the kernel over a local region and computing an inner product. Multiple kernels yield multiple output channels.

\subsection{Graph-Theoretic Interpretation with Regularized Anchor Points}

The convolutional layer can be adapted via a standard layer with regularized anchor points. Let $X \in \mathbb{R}^{1 \times p}$ denote the random variable for the corresponding flattened input. Let $\alpha \in \mathbb{R}^{p \times m}$ represent $m$ convolutional filters, each reshaped and flattened to lie in $\mathbb{R}^p$. Then the convolutional feature map with activation is:
\[
Z = \sigma(X \alpha + \beta), \quad Z \in \mathbb{R}^{n \times m},
\]
where $\beta \in \mathbb{R}^{1 \times m}$ is a bias vector, and $\sigma$ is the elementwise activation function (typically ReLU in CNN). The $j$-th column of $\alpha$, denoted $\alpha_j$, corresponds to the $j$-th filter.

In classical convolution, each filter $\alpha_j$ is nonzero only on a small, spatially localized region $\mathcal{R}_j^c \subseteq \{1, \dots, p\}$. That is:
\[
\mathrm{supp}(\alpha_j) \subseteq \mathcal{R}_j^c, \quad \text{with } |\mathcal{R}_j^c| \ll p.
\]
We thus interpret $\alpha_j$ as a regularized anchor point, localized over the receptive field $\mathcal{R}_j^c$. The output $Z_j = A(X, \alpha_j) = \sigma(X \alpha_j + \beta_j)$ continues the interpretation of each neuron as a pairwise interaction between $X$ and anchor point $\alpha_j$, modulated by the activation $\sigma$.

\subsection{Region Separation for Convolution}

To prove sufficiency for region-separated random variable, it suffices to prove that the convolutional layer, or more precisely under our framework, there exists regularized anchor points such that the transformation is region-separating. An additional patch condition is needed, which is explained in the next subsection. 

\begin{restatable}{theorem}{thmTen}
\label{thm:conv-relu-separation-refined}
Let $X \in \mathbb{R}^p$ be a random variable that is region-separated with respect to $m$ measurable regions $\{\mathcal{R}_1, \dots, \mathcal{R}_m\}$, each is convex and of positive measure.

Suppose that for each $i = 1, \dots, m$, there exists coordinate subsets $\mathcal{Q}_1, \dots, \mathcal{Q}_m \subseteq \{1, \dots, p\}$ and fixed reference patches $x_1, \dots, x_m$ such that:
\[
\forall x \in \mathcal{R}_i, \quad \|x_{\mathcal{Q}_i} - x_i\| =0,
\] 
where for any $i \ne j$, the patterns disagree on their shared patch:
\[
x_i|_{\mathcal{Q}_i \cap \mathcal{Q}_j} \ne x_j|_{\mathcal{Q}_i \cap \mathcal{Q}_j}.
\]

Then there exist sparse vectors $\{\alpha_1, \dots, \alpha_m\} \subset \mathbb{R}^p$ with $\mathrm{supp}(\alpha_i) \subseteq \mathcal{Q}_i$, and biases $\beta \in \mathbb{R}^m$, such that the ReLU-activated convolutional output
\[
Z(x) = \left[ \sigma(\langle x, \alpha_1 \rangle + \beta_1), \dots, \sigma(\langle x, \alpha_m \rangle + \beta_m) \right] \in \mathbb{R}_{\geq 0}^m
\]
is region-separating: for any $x \in \mathcal{R}_i$, $x' \in \mathcal{R}_j$ with $i \ne j$, we have $Z(x) \ne Z(x')$.
\end{restatable}

Next, instead of exact patches, we consider a more general version with a bound in the patch condition.

\begin{restatable}{theorem}{thmEleven}
\label{cor:epsilon-region-separation}
Let $X \in \mathbb{R}^p$ be a region-separated random variable with $m$ disjoint and convex regions $\{\mathcal{R}_1, \dots, \mathcal{R}_m\}$, and suppose there exist coordinate subsets $\mathcal{Q}_1, \dots, \mathcal{Q}_m \subseteq \{1, \dots, p\}$ and fixed reference patches $x_1, \dots, x_m$ such that:
\[
\forall x \in \mathcal{R}_i, \quad \|x_{\mathcal{Q}_i} - x_i\| \le \delta.
\]

Then, using the same sparse filters $\alpha_i$ and biases $\beta_i$ as in Theorem~\ref{thm:conv-relu-separation-refined}, the ReLU output
\[
Z(x) := \left[ \sigma(\langle x, \alpha_1 \rangle + \beta_1), \dots, \sigma(\langle x, \alpha_m \rangle + \beta_m) \right]
\]
satisfies:
\[
\|Z(x) - Z(x')\| \ge \gamma(\delta) > 0 \quad \text{for } x \in \mathcal{R}_i, \ x' \in \mathcal{R}_j,\ i \ne j,
\]
where $\gamma(\delta) \to 0$ as $\delta \to 0$. That is, $Z$ is approximately region-separating, with separation error controlled by $\delta$.
\end{restatable}

\subsection{Sufficiency under the Patch Condition}
We first state an obvious corollary based on the results thus far:
\begin{restatable}{corollary}{corFour}
\label{cor:conv-relu-sufficiency}
Under the assumptions of Theorem~\ref{thm:conv-relu-separation-refined}, the transformation:
\[
T(X) = \sigma(X \alpha + \beta) = Z
\]
is a sufficient statistic for $X$ with respect to any target variable $Y$ measurable with respect to $X$. That is:
\[
P(Y \mid T(X)) = P(Y \mid X).
\]
\end{restatable}

Now, about the additional patch condition---that for any $i \ne j$, the fixed patterns $x_i$ and $x_j$ disagree on their shared support $\mathcal{Q}_i \cap \mathcal{Q}_j$---is essential for ensuring region separation. 

To see this, note that in our construction, each neuron $Z_i(x)$ computes a ReLU-activated inner product between $x$ and a sparse filter $\alpha_i$ supported on $\mathcal{Q}_i$, followed by a bias shift. For the representation $Z(x)$ to be region-separating, we require that $Z(x) \ne Z(x')$ whenever $x$ and $x'$ lie in different regions.

If two distinct regions $\mathcal{R}_i$ and $\mathcal{R}_j$ have identical values on the overlapping coordinates $\mathcal{Q}_i \cap \mathcal{Q}_j$, then it is possible that:
\[
\langle x, \alpha_i \rangle = \langle x', \alpha_i \rangle \quad \text{for } x \in \mathcal{R}_i,\, x' \in \mathcal{R}_j,
\]
which implies $Z_i(x) = Z_i(x')$, defeating the purpose of the neuron as a region detector. The assumption $x_i|_{\mathcal{Q}_i \cap \mathcal{Q}_j} \ne x_j|_{\mathcal{Q}_i \cap \mathcal{Q}_j}$ ensures that the inner products differ, and hence that $Z_i(x) \ne Z_i(x')$ for $i \ne j$.

This assumption matches well with the design of convolutional neural networks (CNNs), particularly for structured inputs like image patches. In images, it is common for each region (e.g., part of a digit, texture, or object) to be characterized by a distinct local pattern over a spatial patch. The index set $\mathcal{Q}_i$ corresponds to the receptive field of a convolutional kernel, and the condition ensures that each region has a distinctive signal over its local patch, which the convolutional filter $\alpha_i$ is designed to detect. Many datasets naturally satisfy this condition:
\begin{itemize}
    \item In digit classification (e.g., MNIST), different digits often contain distinctive local strokes or curves in different parts of the image.
    \item In texture or material recognition, patches exhibit local structure that differs across categories.
    \item In object recognition, local parts (ears, wheels, eyes) appear in roughly consistent locations but differ across object classes.
\end{itemize}

\section{Simulations}
\label{sec:exp}

\subsection{Standard Neural Networks and Region Separation}
We simulate a 3-class classification setting in $\mathbb{R}^3$, where each class consists of 500 points: 400 points are drawn from class-specific Gaussian distributions with means $[10, 0, 0]$, $[0, 10, 0]$, and $[0, 0, 10]$, respectively, and the remaining 100 points per class are drawn from a shared standard Gaussian centered at the origin. While these mixed points are assigned the same class labels during training, they are colored in gray for visualization, to highlight their ambiguity in the input space.

This setting is constructed to illustrate how neural networks preserve information across layers. The input is a region-separated random variable: the three outer regions are class-specific and have conditional probabilities equal to 1 for their respective labels, while the central region (near the origin) corresponds to inputs where $\mathbb{P}(Y \mid X)$ is uniformly distributed across all three classes.

\begin{figure}[h]
	\centering
	\includegraphics[width=0.7\textwidth,trim={0cm 0cm 0cm 0cm},clip]{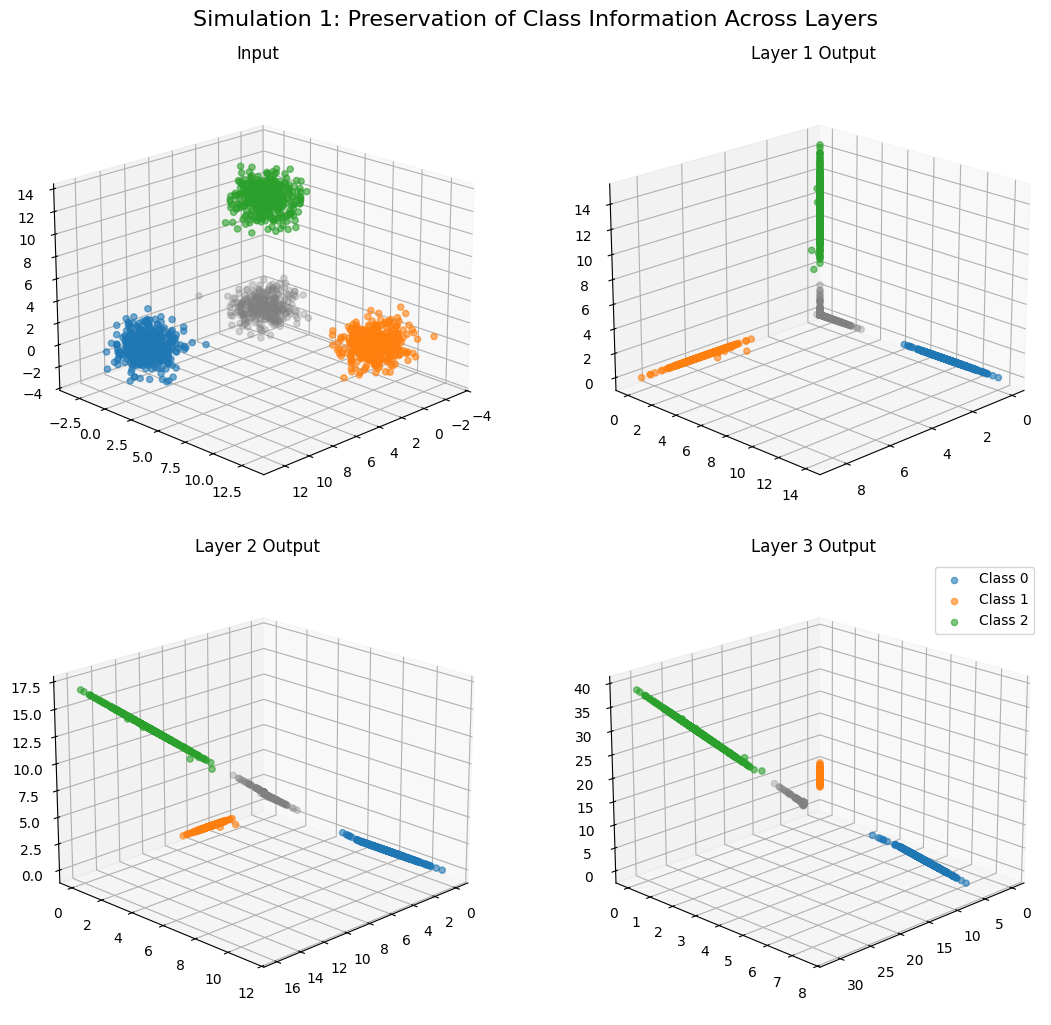}
	\caption{Input data consists of three classes in $\mathbb{R}^3$ with partially overlapping components. Each subplot shows the output representation at a different stage of a 3-layer ReLU network. Class-specific points are shown in blue, orange, and green, while mixed points are plotted in gray. As depth increases, the network maintains class separation for the clean regions and learns representations that preserve sufficient information for classification.}
	\label{fig1}
\end{figure}

We train a standard fully connected neural network with three hidden layers (each with 3 ReLU neurons) and a softmax output layer. Figure~\ref{fig1} visualizes the learned representations in 3D at each layer of the network. The class-specific regions remain clearly separable throughout, while the ambiguous (mixed) points remain clustered near the origin. This experiment demonstrates that when the input distribution is region-separated, neural networks preserve the essential information required for optimal classification, even as they apply nonlinear transformations that distort the input geometry.

It is worth noting that this example is intentionally simple to make the structure and information preservation visually interpretable. In real-world scenarios, the data geometry can be far more complex and high-dimensional, making region separation difficult to visualize. Nonetheless, the key message of the simulation resonates with the theoretical results: if the input distribution is region-separated (or approximately so), neural networks are capable of preserving the optimal classification error through their layers.

\subsection{Convolutional Layers and Patch Condition}

Here, we simulate a 3-class image classification task using synthetic $8 \times 8$ grayscale images. Each image is constructed to contain a class-specific $3 \times 3$ bright patch at one of three distinct spatial locations: top-left, center, or bottom-right, corresponding to class labels 1, 2, and 3 respectively. The rest of the image is filled with low-magnitude Gaussian noise to simulate natural background variability.

This design satisfies the condition from Theorem~\ref{thm:conv-relu-separation-refined}: each class has a distinct pattern on a localized coordinate subset $\mathcal{Q}_i$, while remaining pixels vary but do not interfere with the region-separating structure. It also mimics real-world image classification tasks, where localized discriminative features are embedded in noisy backgrounds.

We train a shallow convolutional neural network with a single convolutional layer of three $3 \times 3$ filters (stride 1, no padding), followed by ReLU activation and global average pooling. The outputs are passed to a softmax classifier.

\begin{figure}[h]
	\centering
	\includegraphics[width=0.6\textwidth,trim={0cm 0cm 0cm 0cm},clip]{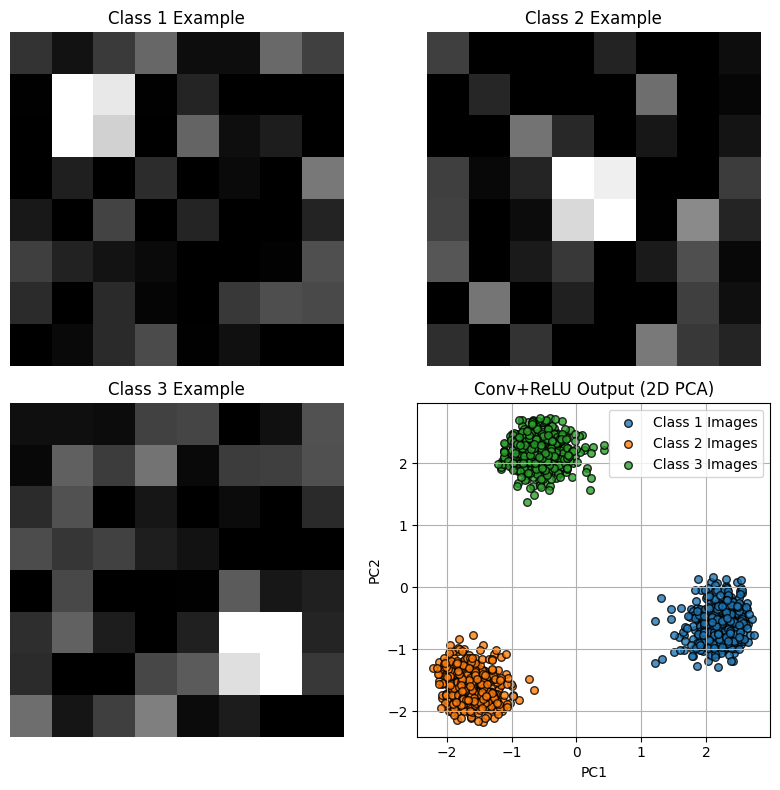}
	\caption{Illustration of region separation in convolutional neural networks. Left: one example image from each of the three classes, where class identity is determined by the location of a bright $3 \times 3$ patch, embedded in noisy background. Right: PCA projection of the convolutional output after ReLU and global pooling. The network successfully separates classes based on local patterns, demonstrating that convolution preserves the sufficient information needed for classification in region-separated input distributions.}
	\label{fig2}
\end{figure}

Figure~\ref{fig2} shows three representative images (one per class) and the 2D PCA projection of the final-layer outputs. The network successfully separates the classes based on the local patch structure, demonstrating that convolutional layers preserve the information needed for classification in region-separated settings. This supports our theoretical claim that convolutional architectures can achieve sufficiency when local region patterns distinguish the classes.

\section{Conclusion}
\label{sec:conclusion}

This paper presents a new graph-based statistical framework for understanding when and how neural network layers preserve information about the target variable $Y$. We approach this from the perspective of statistical sufficiency and graph-like anchor points, formalizing conditions under which neural representations $Z$ retain the conditional distribution $\mathbb{P}(Y \mid X)$. Our key results revolve around two core regimes: sufficiency via injectivity and sufficiency via region separation.



\begin{figure}[h]
	\centering
	\includegraphics[width=0.4\textwidth,trim={0cm 0cm 0cm 0cm},clip]{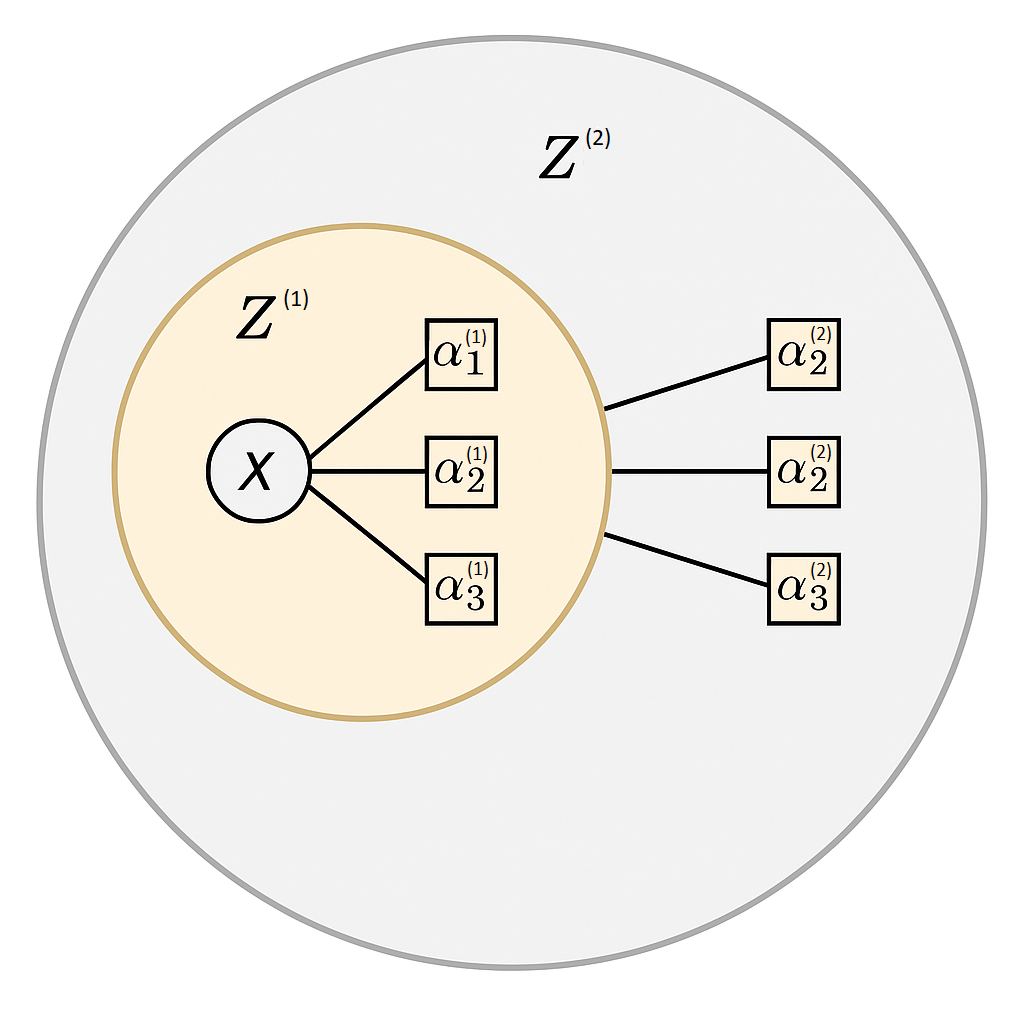}
	\caption{A simplified illustration of a multi-layer neural network viewed as nested graph variables. The input $X$ (center) is transformed into the first-layer representation $Z^{(1)}$ by applying pairwise functions (e.g., inner products) with learned anchor points $\{\alpha_1^{(1)}, \alpha_2^{(1)}, \alpha_3^{(1)},\ldots\}$. The second layer $Z^{(2)}$ is then formed by applying new anchor points $\{\alpha_1^{(2)}, \alpha_2^{(2)}, \alpha_3^{(2)},\ldots\}$ to $Z^{(1)}$. Each layer corresponds to a graph-like variable defined by the pairwise interactions, where sufficiency can be preserved through the nested graphs, i.e., $Prob(Y|X) = Prob(Y|Z^{(1)})=Prob(Y|Z^{(2)})=\ldots$.}
	\label{fig0}
\end{figure}

The injectivity condition is data-agnostic and holds universally given sufficiently large width, whereas region separation is data-adaptive: it requires trained anchor points to specialize to the data distribution, thereby aligning more closely with modern architectures in practice. Table~\ref{table1} summarizes the key differences between the two theoretical paths, while Figure~\ref{fig0} illustrates the overall graph sufficiency perspective of neural networks. In our view, each neural network layer constructs a graph variable by computing pairwise interactions between the input and a set of anchor points. A multi-layer neural network then corresponds to a nested formation of such graph variables, where each graph preserves information from the previous graph. 

\begin{table}[h]
\small
\centering
\caption{Comparison of Injectivity-Based vs. Region-Separated Sufficiency}
\begin{tabularx}{0.9\textwidth}{l|X|X}
\toprule
\textbf{Property} & \textbf{Injectivity-Based} & \textbf{Region-Separated} \\
\midrule
Assumption on $X$ & None & Region-separated distributions \\
\hline
Neuron Width & Must be infinite & Can be finite \\
\hline
Layer Size & Must be finite & Can be finite or infinite \\
\hline
Layer Types & General (e.g., ReLU, sigmoid) & General + Convolutional (with local patches) \\
\hline
Form of Sufficiency & Asymptotic sufficiency wrt neuron width & Exact sufficiency or approximate with bound \\
\hline
Anchor Requirements & Dense in support of $X$ & By construction \\
\hline
Model Training & Holds throughout training under density assumptions & Holds when proper anchor points are learned \\
\bottomrule
\end{tabularx}
\label{table1}
\end{table}

Our work complements but differs fundamentally from classical universal approximation theorems. While universal approximation guarantees that neural networks can approximate any continuous function, our results are distribution-based and focus on whether the internal representations in neural networks retain enough information to permit statistically optimal prediction. This sufficiency-based analysis reveals when and how neural networks can preserve the population-level signal, regardless of the functional relationship. 


Overall, this work provides a new statistical framework to characterize the sufficiency property of neural networks across a range of components. By bridging ideas from graph representations, statistical sufficiency, and modern deep learning, it provides a statistical foundation for understanding when and how neural networks preserve the population-level signal necessary for optimal prediction. We hope this perspective inspires future work in theory-guided architecture design, and deeper exploration for better and more principled development of neural networks.

\bibliographystyle{abbrvnat}
\bibliography{general, shen}

\if1\blind{
\clearpage
\appendix

\bigskip
\begin{center}
{\large\bf APPENDIX}
\end{center}

\section{Proofs}

\lemOne*
\begin{proof}
Since $Z = T(X)$ and $T$ is injective, there exists a measurable inverse function $T^{-1}$ on the range of $T$. Therefore, $X = T^{-1}(Z)$ is a measurable function of $Z$, and hence:
\[
P(Y \mid Z) = P(Y \mid T(X)) = P(Y \mid X).
\]
\end{proof}

\lemTwo*
\begin{proof}
Since $F_\alpha = F_X$, the anchor points $\{\alpha_j\}$ are drawn i.i.d. from the same distribution as $X$. By classical results in probability, with probability one, the empirical set $\{\alpha_j\}_{j=1}^\infty$ becomes dense in the support of $F_X$. This follows from the fact that i.i.d. samples eventually fall into every open set of positive measure, and the separability of compact subsets of $\mathbb{R}^p$ ensures a countable basis to cover the support.
\end{proof}

\lemThree*
\begin{proof}
At $t = 0$, since the $\alpha_j^0$ are drawn i.i.d. from $F_X$, and $\operatorname{supp}(X)$ is compact, it follows from Lemma~\ref{cor:suff-iid-fx} that as $m \to \infty$, the set $\{\alpha_j^0\}$ is dense in $\operatorname{supp}(X)$ with probability 1.

Now fix any sample $\{\alpha_j^0\}_{j=1}^\infty$ such that this denseness holds (which occurs with probability 1). Since the update map $f_t$ is a homeomorphism (by assumption (iii)), and $\{\alpha_j^t\}$ is obtained by applying $f_t$ to $\{\alpha_j^{t-1}\}$, it follows by a standard result in topology that the image of a dense set under a homeomorphism is dense in the image.

Moreover, by assumption (ii), each iterate $\alpha_j^t$ lies in a compact set $\operatorname{supp}(X)^t$, and the updates map one compact support to the next. Therefore, if $\{\alpha_j^{t}\}$ is dense in $\operatorname{supp}(X)^t$, then $\{\alpha_j^{t+1}\} = f_t(\alpha_j^t)$ is dense in $\operatorname{supp}(X)^{t+1}$.

By induction on $t$, the density property is preserved at every step. Since the initial density holds with probability 1 as $m \to \infty$, so does the density of $\{\alpha_j^t\}$ in $\operatorname{supp}(X)^t$ for all $t$.
\end{proof}

\thmOne*
\begin{proof}
Let $x_1 \ne x_2$ in $\operatorname{supp}(X)$. By assumption (i), there exists $\alpha^*$ such that $A(x_1, \alpha^*) \ne A(x_2, \alpha^*)$. Since $\{\alpha_j\}$ is dense and $A(x, \cdot)$ is continuous, we can find $\alpha_j$ arbitrarily close to $\alpha^*$ such that $A(x_1, \alpha_j) \ne A(x_2, \alpha_j)$. Therefore, for sufficiently large $m$, the vectors $Z_m(x_1)$ and $Z_m(x_2)$ differ — i.e., $Z_m$ is injective on $\operatorname{supp}(X)$ as $m \to \infty$.

Since $Z_m$ becomes injective on $\operatorname{supp}(X)$ as $m \to \infty$, the map $X \mapsto Z_m$ becomes invertible in the limit, and $X$ becomes a measurable function of $Z_m$. As a result, the conditional distribution $\mathbb{P}(Y \mid Z_m)$ converges to the true conditional $\mathbb{P}(Y \mid X)$.

Moreover, since $Y$ has finite second moment, the conditional probability function $\mathbb{P}(Y \mid X)$ is square-integrable, and so:
\[
\lim_{m \to \infty} \left\| \mathbb{P}(Y \mid Z_m) - \mathbb{P}(Y \mid X) \right\|_{L^2(F_X)} = 0.
\]
Thus, the representation $Z_m$ is asymptotically sufficient for $X$ with respect to $Y$.
\end{proof}

\thmTwo*
\begin{proof}
We apply Theorem~\ref{thm:asymp-suff-dense} by verifying that each pairwise function $A(x, \alpha)$ satisfies point separation, i.e., for any $x_1 \ne x_2 \in \operatorname{supp}(X)$, there exists $\alpha$ such that $A(x_1, \alpha) \ne A(x_2, \alpha)$, as well as continuity in $\alpha$.

(a) Inner product:
For $x_1 \ne x_2$, we have $x_1 - x_2 \ne 0$, so there exists $\alpha$ such that $\langle x_1, \alpha \rangle \ne \langle x_2, \alpha \rangle$. Since inner product is continuous and bounded on compact sets, both conditions hold.

(b) Euclidean distance:
The function $\alpha \mapsto \|x - \alpha\|$ is continuous and strictly convex, and $\|x_1 - \alpha\| = \|x_2 - \alpha\|$ defines a hypersurface in $\mathbb{R}^p$. Thus, for almost every $\alpha$, $A(x_1, \alpha) \ne A(x_2, \alpha)$, and the dense sequence $\{\alpha_j\}$ suffices to ensure point-separation. Continuity and integrability hold on compact domains.

(c) Squared distance:
Since $A(x, \alpha) = \|x - \alpha\|^2$ is strictly increasing in the distance and smooth in $\alpha$, point-separation and continuity follow from case (b). Square-integrability is preserved.

(d) Universal kernel:
A continuous universal kernel $k$ on compact $\operatorname{supp}(X)$ defines an injective embedding $x \mapsto k(x, \cdot)$ into its RKHS $\mathcal{H}_k$. Since $\{\alpha_j\}$ is dense and $k$ is continuous in both arguments, the evaluations $\{k(x, \alpha_j)\}$ uniquely determine $x$ via the kernel map. Thus, the conditions of Theorem~\ref{thm:asymp-suff-dense} are satisfied.

(e) Cosine similarity:
Define $\tilde{X} := X / \|X\|$ on $\operatorname{supp}(X) \setminus \{0\}$. Under the assumption $Y \perp \|X\| \mid \tilde{X}$, we have:
\[
\mathbb{P}(Y \mid X) = \mathbb{P}(Y \mid \tilde{X}),
\]
so sufficiency for $\tilde{X}$ implies sufficiency for $Y$. Now note:
\[
A(X, \alpha) = \frac{\langle X, \alpha \rangle}{\|X\| \|\alpha\|} = \langle \tilde{X}, \tilde{\alpha} \rangle,
\]
with $\tilde{\alpha} = \alpha / \|\alpha\|$. On the unit sphere, inner products again separate points (as in case (a)), and continuity holds. Thus, cosine similarity yields an asymptotically sufficient representation for $\tilde{X}$.

If the norm $\|X\|$ carries additional information about $Y$, then the augmented representation $[Z_m; \|X\|]$ fully reconstructs $X$ asymptotically and is therefore sufficient.
\end{proof}

\corOne*
\begin{proof}
Since addition of a fixed vector is a bijective affine transformation on $\mathbb{R}^m$, injectivity of the map $X \mapsto Z$ implies injectivity of $X \mapsto \tilde{Z}$. That is,
\[
X_1 \ne X_2 \implies Z(X_1) \ne Z(X_2) \implies \tilde{Z}(X_1) \ne \tilde{Z}(X_2).
\]
\end{proof}

\thmThree*
\begin{proof}
By Theorem~\ref{thm:asymp-suff-dense}, if $\{\alpha_j\}$ is dense and $A(x, \alpha)$ separates points, then the map $X \mapsto Z$ is injective as $m \to \infty$. Adding any fixed bias $\beta \in \mathbb{R}^m$ preserves injectivity, so $X \mapsto Z + \beta$ remains injective in the limit.

Now consider $\tilde{Z} = \sigma(Z + \beta)$. The ReLU function is continuous, monotonic, and injective on $[0, \infty)$, but non-invertible on $\mathbb{R}$ due to its flat zero region on $(-\infty, 0]$.

Suppose for contradiction that $x_1 \ne x_2$ in $\operatorname{supp}(X)$ yield $\tilde{Z}(x_1) = \tilde{Z}(x_2)$. Then $Z(x_1) + \beta$ and $Z(x_2) + \beta$ differ, but their ReLU outputs agree. This can only happen if the differences lie entirely in negative coordinates that are zeroed out — i.e., all nonzero differences lie in coordinates where both values are negative.

But this contradicts the injectivity of $Z + \beta$: if all coordinates where $(Z + \beta)(x_1) \ne (Z + \beta)(x_2)$ are clipped to zero by ReLU, then $Z(x_1) + \beta = Z(x_2) + \beta$ on the support of nonzero output, which implies $x_1 = x_2$. Therefore, such $x_1 \ne x_2$ cannot exist.

Thus, $\tilde{Z}$ remains injective in the limit. By the same reasoning as in Theorem~\ref{thm:asymp-suff-dense}, this implies that $\mathbb{P}(Y \mid \tilde{Z}) \to \mathbb{P}(Y \mid X)$ in $L^2(F_X)$ norm, completing the proof.
\end{proof}

\thmFour*
\begin{proof}
We prove the result by induction on the layer index $\ell$.

Base case ($\ell = 1$):
Since the anchors $\{\alpha_j^{(1)}\}$ are dense in $\operatorname{supp}(X)$ with probability 1, and $A^{(1)}(x, \alpha)$ separates points and is continuous in $\alpha$, Theorem~\ref{thm:asymp-suff-dense} implies that $Z^{(1)}$ becomes an injective function of $X$ as $m^{(1)} \to \infty$. Therefore,
\[
\lim_{m^{(1)} \to \infty} \left\| \mathbb{P}(Y \mid Z^{(1)}) - \mathbb{P}(Y \mid X) \right\|_{L^2(F_X)} = 0.
\]
Moreover, since $Z^{(1)}$ is a continuous function of $X$, it is supported on a compact set.

Inductive step:
Assume the result holds up to layer $\ell - 1$, so that $Z^{(\ell-1)}$ is injective in the limit and supported on a compact set. By assumption, $\{\alpha_j^{(\ell)}\}$ remains dense in this support, and $A^{(\ell)}$ satisfies the same continuity and separation conditions. Applying Theorem~\ref{thm:asymp-suff-dense} again gives:
\[
\lim_{m^{(\ell)} \to \infty} \left\| \mathbb{P}(Y \mid Z^{(\ell)}) - \mathbb{P}(Y \mid Z^{(\ell-1)}) \right\|_{L^2(F_X)} = 0.
\]

Final conclusion:  
Since $L$ is fixed and finite, we apply the triangle inequality:
\[
\left\| \mathbb{P}(Y \mid Z^{(L)}) - \mathbb{P}(Y \mid X) \right\|_{L^2(F_X)} 
\le \sum_{\ell=1}^L \left\| \mathbb{P}(Y \mid Z^{(\ell)}) - \mathbb{P}(Y \mid Z^{(\ell-1)}) \right\|_{L^2(F_X)},
\]
and each term tends to zero as $m^{(\ell)} \to \infty$. Taking the joint limit over all widths, the total deviation vanishes, completing the proof.
\end{proof}

\lemFive*
\begin{proof}
Since $X$ takes on only $m$ discrete values, we may identify the support of $X$ with the finite set $\{x_{1}, \dots, x_{m}\}$. The assumption that $T$ is injective over this set implies that for each $x_{i}$, there is a unique $z_i = T(x_{i})$. Thus, $T$ induces a bijection between $\{x_{i}\}$ and $\{z_i\}$.

Let $Z = T(X)$ denote the transformed variable. Since $T$ is injective on $\operatorname{supp}(X)$, there exists an inverse mapping $T^{-1}$ defined on $\{z_1, \dots, z_m\}$ such that $T^{-1}(T(x_{i})) = x_{i}$. It follows that:
\[
P(Y \mid Z) = P(Y \mid X),
\]
almost surely, which proves that $Z$ is a sufficient statistic for $X$ with respect to $Y$.
\end{proof}

\thmFive*
\begin{proof}
Let $Z = T(X)$, and assume $T$ is region-separating, i.e., the map $x \mapsto Z$ maps all $x \in \mathcal{R}_i$ to some fixed value $z_i$, distinct from $z_j$ for $j \ne i$. This induces a deterministic partition of $Z$ into $\{z_1, \dots, z_m\}$, each corresponding to a region $\mathcal{R}_i$.

Since $P(Y \mid X = x) = q_i$ for all $x \in \mathcal{R}_i$, and $Z$ uniquely identifies the region $x$ belongs to, we have:
\[
P(Y \mid Z = z_i) = q_i = P(Y \mid X = x), \quad \forall x \in \mathcal{R}_i.
\]
Therefore,
\[
P(Y \mid Z) = P(Y \mid X)
\]
almost surely, which proves sufficiency.
\end{proof}

\thmSix*
\begin{proof}
Since $f \in L^1(F_X)$ and $X$ is supported on the compact set, $f$ is measurable on the support, by Lusin’s theorem, for any $\delta > 0$, there exists a compact set $\mathcal{X}$ such that:
\[
F_X(\mathcal{X}) \geq 1 - \delta, \quad \text{and} \quad f \text{ is continuous on } \mathcal{X}.
\]

Since $f$ is continuous on compact $\mathcal{X}$, it is uniformly continuous. Thus, for any $\epsilon > 0$, there exists $\eta > 0$ such that:
\[
\|x - x'\| < \eta \quad \Rightarrow \quad \|f(x) - f(x')\|_1 < \epsilon, \quad \forall x, x' \in \mathcal{X}.
\]

We can cover $\mathcal{X}$ by a finite number of balls $\{B_1, \dots, B_m\}$ of radius at most $\eta$. Define regions $\mathcal{R}_i := B_i \cap \mathcal{X}$ for $i = 1, \dots, m$. These form a measurable cover of $\mathcal{X}$, and we may refine them into a disjoint measurable partition (e.g., by subtracting overlaps or assigning boundaries arbitrarily). By construction, for all $x, x' \in \mathcal{R}_i$, $\|f(x) - f(x')\|_1 \leq \epsilon$.

Then, let $\mathcal{R}_0 := \mathrm{supp}(X) \setminus \mathcal{X}$. Since $F_X(\mathcal{R}_0) \leq \delta$, we may assign $\mathcal{R}_0$ arbitrarily (e.g., as an $(m+1)$-th region or absorb it into the nearest $\mathcal{R}_i$). For integrated approximations, its effect is bounded:
\[
\int_{\mathcal{R}_0} \|f(x) - \bar{f}(x)\|_1 \, dF_X(x) \leq 2\delta,
\]
for any constant $\bar{f}$ over $\mathcal{R}_0$.

Choosing $\delta < \epsilon / 2$, we obtain a finite measurable partition $\{\mathcal{R}_1, \dots, \mathcal{R}_m\}$ such that $\|f(x) - f(x')\|_1 \leq \epsilon$ for all $x, x'$ in any region. Hence, $X$ is $\epsilon$-region-separated with $m$ regions, as desired.
\end{proof}

\thmSeven*
\begin{proof}
Since $T$ is region-separating, it maps each region $\mathcal{R}_i$ to a unique point $z_i = T(x)$ for all $x \in \mathcal{R}_i$. Hence, the support of $Z = T(X)$ is exactly $\{z_1, \dots, z_m\}$.

For each $i$, the event $\{Z = z_i\}$ corresponds exactly to $\{X \in \mathcal{R}_i\}$. Therefore, the conditional distribution of $Y$ given $Z = z_i$ is:
\[
\mathbb{P}(Y \mid Z = z_i) = \mathbb{E}[ \mathbb{P}(Y \mid X) \mid X \in \mathcal{R}_i ].
\]

Now fix any $x \in \mathcal{R}_i$. Using the triangle inequality and the definition of $\epsilon$-region separation:
\[
\left\| \mathbb{P}(Y \mid Z = z_i) - \mathbb{P}(Y \mid x) \right\|_1
= \left\| \mathbb{E}_{x' \in \mathcal{R}_i} [\mathbb{P}(Y \mid x')] - \mathbb{P}(Y \mid x) \right\|_1
\leq \epsilon.
\]

Thus, the conditional distribution $\mathbb{P}(Y \mid Z = z_i)$ approximates $\mathbb{P}(Y \mid x)$ uniformly within each region $\mathcal{R}_i$ up to total variation $\epsilon$.

Hence, $Z$ is $\epsilon$-region-separated with respect to the induced partition $\{z_1, \dots, z_m\}$.
\end{proof}

\thmEight*
\begin{proof}
Assume each region $\mathcal{R}_i \subset \mathbb{R}^p$ is compact, convex, and has nonzero interior. For each region $\mathcal{R}_i$, choose a representative point $c_i \in \mathcal{R}_i$. Define:
\[
\alpha_j := c_j, \quad b_j := -\|c_j\|^2, \quad \text{for } j = 1, \dots, m.
\]
Then the map
\[
Z(x) := [\langle x, \alpha_1 \rangle + b_1, \dots, \langle x, \alpha_m \rangle + b_m]
\]
satisfies:
\[
Z(c_i) = [\langle c_i, c_1 \rangle - \|c_1\|^2, \dots, \langle c_i, c_m \rangle - \|c_m\|^2].
\]

We claim that for a generic choice of representatives $\{c_1, \dots, c_m\}$, the vectors $\{Z(c_i)\}$ are distinct. This follows because: if $Z(c_i) = Z(c_j)$ for $i \ne j$, then $c_i - c_j$ must lie in the orthogonal complement of the span of $\{c_1, \dots, c_m\}$, which is a measure-zero event. Since each region $\mathcal{R}_i$ has positive measure and contains interior points, such collisions can be avoided by re-selecting $c_i$ from a full-measure subset. Thus, we may choose the $c_i$ so that $Z(c_i) \ne Z(c_j)$ for all $i \ne j$.

Now, because $Z$ is affine and continuous, it maps each small neighborhood around $c_i$ in $\mathcal{R}_i$ to an open set around $Z(c_i)$ in $\mathbb{R}^m$. Since each region is convex, for each $c_j$, there exists a hyperplane, namely, $\langle x, c_j \rangle = \|c_j\|^2$, that separates a neighborhood of $c_j$ within $\mathcal{R}_j$ from the other regions. This guarantees that for sufficiently small neighborhoods around $c_i$ and $c_j$, the map $Z$ produces distinct outputs.

Hence, for almost every $x \in \mathcal{R}_i$ and $x' \in \mathcal{R}_j$ with $i \ne j$, we have $Z(x) \ne Z(x')$. Therefore, $Z$ separates the regions almost surely.
\end{proof}

\thmNine*
\begin{proof}
For each region $\mathcal{R}_i$, select a representative point $c_i \in \mathcal{R}_i$. Define $\alpha_j := c_j$ and $b_j := -\|c_j\|^2$ for $j = 1, \dots, m$. Then each coordinate of the representation is:
\[
Z_j(x) = \max(0, \langle x, c_j \rangle - \|c_j\|^2).
\]

By construction: $Z_j(x) > 0$ if $\langle x, c_j \rangle > \|c_j\|^2$, and $Z_j(x) = 0$ if $\langle x, c_j \rangle \le \|c_j\|^2$.

In particular, for $x = c_i$:
\[
Z_i(c_i) = \max(0, \|c_i\|^2 - \|c_i\|^2) = 0, \quad
Z_j(c_i) = \max(0, \langle c_i, c_j \rangle - \|c_j\|^2) \le 0 \text{ for } j \ne i.
\]

To ensure that $Z(c_i) \ne Z(c_j)$ for $i \ne j$, we may reselect $c_i$ from $\mathcal{R}_i$ if needed — since each region has positive measure, this is always possible.

By continuity of $Z(x)$ and the fact that each region has nonzero interior and convex, the pattern $Z(x) \approx Z(c_i)$ holds for almost all $x \in \mathcal{R}_i$, and $Z(x) \ne Z(x')$ for $x' \in \mathcal{R}_j$, $j \ne i$, except possibly on boundaries of measure zero.

Hence, $Z$ separates regions almost surely.
\end{proof}

\corTwelve*
\begin{proof}
Since $T$ is region-separating, we have $T(x) = z_i$ for all $x \in \mathcal{R}_i$. Define $Z = T(X)$. Then for any $z_i = T(x)$, and any $x \in \mathcal{R}_i$:
\[
P(Y \mid Z = z_i) = \mathbb{E}_{X \mid Z=z_i}[P(Y \mid X)].
\]

Since $P(Y \mid X)$ varies by at most $\epsilon$ within each $\mathcal{R}_i$, and $Z$ is constant on $\mathcal{R}_i$, it follows that:
\[
\| P(Y \mid Z = z_i) - P(Y \mid X = x) \|_1 \leq \epsilon.
\]
This holds for all $x$ in the same region, proving the bound.
\end{proof}

\thmTwelve*
\begin{proof}
(1) Regression:
Let $\mu_X := \mathbb{E}[Y \mid X]$ and $\mu_Z := \mathbb{E}[Y \mid Z]$. It follows that
\begin{align*}
\operatorname{MSE}^*(Z) 
&= \mathbb{E}[(Y - \mu_X + \mu_X - \mu_Z)^2] \\
&= \mathbb{E}[(Y - \mu_X)^2] 
+ \mathbb{E}[(\mu_X - \mu_Z)^2] 
+ 2\mathbb{E}[(Y - \mu_X)(\mu_X - \mu_Z)] \\
&= \operatorname{MSE}^*(X) + \mathbb{E}[(\mu_X - \mu_Z)^2] + 2\mathbb{E}[(Y - \mu_X)(\mu_X - \mu_Z)].
\end{align*}

Since $Y \in [y_1, y_2]$ and $\|P(Y \mid Z) - P(Y \mid X)\|_1 \leq \epsilon$, it follows that
\[
|\mu_X - \mu_Z| \leq \epsilon (y_2-y_1) \quad \text{(pointwise)}, \quad \text{and} \quad
\operatorname{Var}(Y \mid X) \leq \frac{(y_2-y_1)^2}{4}.
\]

It follows that $\mathbb{E}[(\mu_X - \mu_Z)^2] \leq \epsilon^2 (y_2-y_1)^2$, and 
\begin{align*}
|\mathbb{E}[(Y - \mu_X)(\mu_X - \mu_Z)]| &\leq \sqrt{\mathbb{E}[(Y - \mu_X)^2]} \cdot \sqrt{\mathbb{E}[(\mu_X - \mu_Z)^2]} \\
&\leq \frac{(y_2-y_1)}{2} \cdot \epsilon (y_2-y_1) = \frac{1}{2} \epsilon (y_2-y_1)^2
\end{align*}

Putting it all together:
\[
|\operatorname{MSE}^*(Z) - \operatorname{MSE}^*(X)| \leq \epsilon^2 (y_2 - y_1)^2 + \frac{\epsilon}{2}(y_2 - y_1)^2.
\]

(2) Classification:
For classification, the Bayes-optimal classification error is:
\[
\operatorname{Err}^*(X) = 1 - \mathbb{E}_X \left[ \max_y P(Y = y \mid X) \right], \quad 
\operatorname{Err}^*(Z) = 1 - \mathbb{E}_Z \left[ \max_y P(Y = y \mid Z) \right].
\]

Since $\|P(Y \mid Z) - P(Y \mid X)\|_1 \leq \epsilon$ for all $Y$, this holds for the maximum as well. Applying the same bound pointwise and then integrating gives:
\[
\left| \operatorname{Err}^*(Z) - \operatorname{Err}^*(X) \right| \leq \epsilon.
\]
\end{proof}

\thmTen*
\begin{proof}
For each $i = 1, \dots, m$, define a sparse vector $\alpha_i \in \mathbb{R}^p$ supported on $\mathcal{Q}_i$:
\[
\alpha_i[k] := 
\begin{cases}
x_i[k], & \text{if } k \in \mathcal{Q}_i, \\
0, & \text{otherwise}.
\end{cases}
\]
Let $\beta_i := -\frac{1}{2} \|x_i\|^2$.

Now, for any $x \in \mathcal{R}_i$, we have $x_{\mathcal{Q}_i} = x_i$, so:
\[
\langle x, \alpha_i \rangle = \langle x_i, x_i \rangle = \|x_i\|^2,
\]
and thus:
\[
Z_i(x) = \sigma(\langle x, \alpha_i \rangle + \beta_i) = \sigma\left(\frac{1}{2} \|x_i\|^2\right) > 0.
\]

For $x' \in \mathcal{R}_j$ with $j \ne i$, the inner product becomes:
\[
\langle x', \alpha_i \rangle = \langle x'_{\mathcal{Q}_i}, x_i \rangle.
\]
Since $x'_{\mathcal{Q}_j} = x_j$ and $x_i \ne x_j$ on $\mathcal{Q}_i \cap \mathcal{Q}_j$, there exists some coordinate $k \in \mathcal{Q}_i$ where $x'_{\mathcal{Q}_i}[k] \ne x_i[k]$. Therefore:
\[
\langle x'_{\mathcal{Q}_i}, x_i \rangle < \|x_i\|^2,
\]
and hence:
\[
Z_i(x') = \sigma\left(\langle x', \alpha_i \rangle + \beta_i\right) < \sigma\left(\|x_i\|^2 - \frac{1}{2} \|x_i\|^2\right) = \frac{1}{2} \|x_i\|^2.
\]
If the inner product gap is large enough (e.g., due to orthogonal or contrasting values), we can make $Z_i(x') = 0$ by design.

Therefore, $Z_i(x) > 0$ for all $x \in \mathcal{R}_i$, and $Z_i(x') = 0$ for $x' \in \mathcal{R}_j$, $j \ne i$. Thus, the vector $Z(x)$ is uniquely activated on region $\mathcal{R}_i$ and zero elsewhere.
\end{proof}

\thmEleven*
\begin{proof}
Let $x \in \mathcal{R}_i$, $x' \in \mathcal{R}_j$ with $i \ne j$. By assumption,
\[
\|x_{\mathcal{Q}_i} - x_i\| \le \delta, \quad \|x'_{\mathcal{Q}_j} - x_j\| \le \delta.
\]

Define $\alpha_i \in \mathbb{R}^p$ by zero-padding $x_i$ outside $\mathcal{Q}_i$, and $\beta_i := -\frac{1}{2} \|x_i\|^2$. Then:
\[
\langle x, \alpha_i \rangle = \langle x_{\mathcal{Q}_i}, x_i \rangle = \|x_i\|^2 + \langle x_{\mathcal{Q}_i} - x_i, x_i \rangle,
\]
so
\[
|\langle x, \alpha_i \rangle - \|x_i\|^2| \le \|x_{\mathcal{Q}_i} - x_i\| \cdot \|x_i\| \le \delta \|x_i\|.
\]

Thus,
\[
Z_i(x) = \sigma(\langle x, \alpha_i \rangle + \beta_i) \ge \frac{1}{2} \|x_i\|^2 - \delta \|x_i\|.
\]

Now consider $Z_i(x')$ for $x' \in \mathcal{R}_j$, $j \ne i$. We estimate:
\[
\langle x', \alpha_i \rangle = \langle x'_{\mathcal{Q}_i}, x_i \rangle = \langle x_j|_{\mathcal{Q}_i \cap \mathcal{Q}_j}, x_i \rangle + \text{error},
\]
where the error is at most $\delta \|x_i\|$. By assumption, $x_i \ne x_j$ on $\mathcal{Q}_i \cap \mathcal{Q}_j$, so:
\[
\langle x'_{\mathcal{Q}_i}, x_i \rangle \le \|x_i\|^2 - \Delta \quad \text{for some } \Delta > 0.
\]
Hence:
\[
Z_i(x') \le \max\left(0, \|x_i\|^2 - \Delta + \beta_i + \delta \|x_i\|\right) = \max\left(0, \frac{1}{2} \|x_i\|^2 - \Delta + \delta \|x_i\|\right).
\]

Now define the separation margin:
\[
\gamma(\delta) := \left[ \frac{1}{2} \|x_i\|^2 - \delta \|x_i\| \right] - \left[ \frac{1}{2} \|x_i\|^2 - \Delta + \delta \|x_i\| \right] = \Delta - 2 \delta \|x_i\|.
\]

So as long as $\delta < \Delta / (2 \|x_i\|)$, we have $\gamma(\delta) > 0$, and:
\[
|Z_i(x) - Z_i(x')| \ge \gamma(\delta).
\]

Since this holds for some $i$ depending on $(\mathcal{R}_i, \mathcal{R}_j)$, we get:
\[
\|Z(x) - Z(x')\| \ge \gamma(\delta),
\]
which completes the proof.
\end{proof}

\corFour*
\begin{proof}
Let $T(X) = Z = \sigma(X \alpha + \beta)$ be the transformation constructed in Theorem~\ref{thm:conv-relu-separation-refined}. For each region $\mathcal{R}_i$, $Z(x)$ takes a unique value (e.g., a one-hot or sparse vector with nonzero only in position $i$). Therefore, $Z(x)$ uniquely determines the region $\mathcal{R}_i$ to which $x$ belongs.

Since $P(Y \mid X)$ is constant on each $\mathcal{R}_i$ (because the regions partition $\operatorname{supp}(X)$), and $Z$ identifies the region, we have:
\[
P(Y \mid Z) = P(Y \mid X).
\]
Hence, $Z$ is a sufficient statistic for $X$ with respect to $Y$.
\end{proof}
} \fi

\end{document}